\documentclass[runningheads]{llncs}
\usepackage[T1]{fontenc}
\usepackage{cite}
\usepackage{amsmath,amssymb,amsfonts}
\usepackage{algorithmic}
\usepackage{graphicx}
\usepackage{textcomp}
\usepackage{xcolor}
\usepackage{romannum}
\usepackage{algorithmic}
\usepackage[ruled,vlined,linesnumbered]{algorithm2e}
\usepackage{float}
\usepackage{subfigure}
\usepackage[colorinlistoftodos]{todonotes}
\newcommand{\GreedyTensile}{{\it Greedy Tensile }}

\begin{document}
%
\title{Knowledge-Guided Machine Learning for Stabilizing Near-Shortest Path Routing}

\titlerunning{Knowledge-Guided Machine Learning for APNSP}
%
\author{Yung-Fu Chen\inst{1} \and
Sen Lin\inst{2}\and
Anish Arora\inst{1}}
\authorrunning{Y. Chen et al.}
%
\institute{The Ohio State University, Columbus, OH 43210, USA \\ \email{\{chen.6655, arora.9\}@osu.edu}\and
University of Houston, Houston, Texas 77204, USA \\
\email{slin50@central.uh.edu}
}
\maketitle              
\begin{abstract}
We propose a simple algorithm that needs only a few data samples from a single graph for learning local routing policies that generalize across a rich class of geometric random graphs in Euclidean metric spaces. We thus solve the all-pairs near-shortest path problem by training deep neural networks (DNNs) that let each graph node efficiently and scalably route (i.e., forward) packets by considering only the node's state and the state of the neighboring nodes. Our algorithm design exploits network domain knowledge in the selection of input features and design of the policy function  for learning an approximately optimal policy. Domain knowledge also provides theoretical assurance that the choice of a ``seed graph'' and its node data sampling suffices for generalizable learning. 
Remarkably, one of these DNNs we train ---using distance-to-destination as the only input feature--- learns a policy that exactly matches the well-known Greedy Forwarding policy, which forwards packets to the neighbor with the shortest distance to the destination. We also learn a new policy, which we call \GreedyTensile routing ---using both distance-to-destination and node stretch as the input features--- that almost always outperforms greedy forwarding. We demonstrate the explainability and ultra-low latency run-time operation of Greedy Tensile routing by symbolically interpreting its DNN in low-complexity terms of two linear actions.
\end{abstract}
\vspace*{-10mm}
\section{Introduction}
\vspace*{-2mm}

There has been considerable interest in machine learning to mimic the human ability of learning new concepts from just a few instances. While formal frameworks for machine learning, such as the language-identification-in-the-limit framework \cite{gold1967language} and probably approximately correct (PAC) learning \cite{valiant1984theory}, have yet to match the high data efficiency of human learning, few-shot (or one-shot) learning methods \cite{muggleton1996learning, wang2020generalizing, muggleton2023hypothesizing} are attempting to bridge the gap by using prior knowledge to rapidly generalize to new instances given training on only a small number of samples with supervised information.

In this paper, we explore the learnability of policies that generalize in the domain of graph routing, where a one-size-fits-all solution based on local or global network states is challenging. Routing using global network states tends to be inefficient in time and communication when the class of graphs has strict scalability limits for network capacity \cite{xue2006scaling} or has significant graph dynamics \cite{hekmat2004interference, grossglauser2002mobility}. Manually designed algorithms and heuristics using local or regional network states often cater to particular network conditions and come with tradeoffs of complexity, scalability, and generalizability across diverse graphs. Many machine learned algorithms in this space incur relatively high computational complexity during training \cite{reis2019deep}, have high overhead at run-time that limits their use in graphs with high dynamics, or are applicable only for small-scale graphs or graphs of limited types (e.g., high-density graphs). Our work focuses attention on answering the following question: 
\vspace*{-2mm}
\begin{quote}
\noindent\emph{Can we design a sample-efficient machine learning algorithm for graph routing based on local information that addresses scalability, generalizability, and complexity issues all at once?}  
\end{quote}

\noindent 

We answer this question in the affirmative for the all-pairs near-shortest path (APNSP) problem over the class of uniform random graphs in Euclidean metric spaces. It is well known that uniform random graphs in Euclidean metric spaces can represent the topologies inherent in wireless networks;
the policies we learn are thus applicable to real-world wireless networks. Our key insight is that ---in contrast to pure black-box approaches--- {\em exploiting domain knowledge in various ways --- input feature selection, policy design, sample selection --- enables learning of near-optimal, low-complexity routing from only a few samples that generalizes to (almost) all graphs in this class and adapts quickly to network dynamics.}

\vspace*{2mm}
\noindent
{\bf Near-optimal, low-complexity routing}. To motivate our focus on routing using only local information, we recall that approaches to solving the APNSP problem fall into two categories: global information and local information. Policies using global information encode the entire network state into graph embeddings \cite{narayanan2017graph2vec} and find optimal paths, whereas policies using local information need only node embeddings \cite{grover2016node2vec} or the coordinates of the neighbors and the destination to predict the next forwarder on the optimal path. The complexity (in time and space) resulting from the latter is inherently better than that of the former. However, typically, the latter comes with the penalty of sub-optimality of the chosen path relative to the optimal (i.e., shortest) path. While this holds for both manual designs and machine learned algorithms, we show that by exploiting domain knowledge with respect to bounded forwarding, it is possible to learn a routing policy that achieves near-optimality in (almost) all graphs in the chosen class.

\vspace*{2mm}
\noindent
{\bf Generalizability}. We also develop a domain theory for achieving efficient learning that generalizes over the class of graphs.  The theory is based on a condition under which the ranking of neighboring forwarders based on local information matches that based on globally optimal information. We empirically validate the condition by demonstrating a strong correlation between local ranking and global ranking metrics for the class of graphs. The theory then implies the APNSP objective can be realized with high probability by training a deep neural network (DNN) that characterizes the local ranking metric of each neighbor as a potential forwarder. Moreover, it implies the DNN policy can generalize even if it is trained from only a few data samples chosen from a single ``seed'' graph. The theory also guides our selection of seed graphs as well as corresponding training data and is corroborated by empirical validation of our learned routing solutions.

\vspace*{2mm}
\noindent
{\bf Quick adaptation to network dynamics}. By virtue of its generalizability to all graphs in the class, the same DNN can be used without any retraining when the network changes. Moreover, since the DNN uses only local information, route adaptation to dynamics is quick. In fact, the policy is inherently self-stabilizing.

In this paper, we develop knowledge-guided learning as follows: We model the APNSP problem as a Markov decision process (MDP) and use a DNN architecture to learn a ``single-copy'' local routing policy based on the features of distance-to-destination and node stretch. At each routing node and at each time, the DNN only considers the states from that node and one of its neighbors to predict a local metric (a $Q$-value) for routing. Routing thus uses a single neighbor for which the $Q$-value is the largest. 

The approach yields a routing policy that we call \GreedyTensile routing. \GreedyTensile is a light-weight routing policy for the chosen class of graphs, in following ways: (a) Low complexity: It is rapidly learned from a small dataset that is easily collected from a single ``seed'' graph; (b) Generalizability: It can be used on all nodes of a graph, and is able to generalize across almost all graphs in the class without additional training on the target networks; and (c) Scalability: Its routing decision only depends on the local network state, i.e., the state of the node and its one-hop neighbor nodes, and thus is much more efficient than routing using global states and can be used even when the topology changes.

The main contributions and findings of this paper are summarized as follows: 
\begin{enumerate}
    \item Generalization from few-shot learning from a single graph is feasible for APNSP and theoretically assured by domain knowledge. 
    \item Domain knowledge also guides the selection of input features and training samples to increase the training efficiency and testing accuracy. 
    \item Learning from a single graph using only a distance-to-destination metric matches the well-known greedy forwarding routing. 
    \item Learning from a single graph using both distance-to-destination and node stretch relative to a given origin-destination node pair yields a new policy, \GreedyTensile routing, that achieves even better generalized APNSP routing over Euclidean random graphs.
    \item The \GreedyTensile DNN can be symbolically interpreted in a low-complexity fashion---it is approximated by a policy with two linear actions. 
    \item Reinforcement learning from a single graph achieves comparable generalization performance for APNSP.
\end{enumerate}

\vspace*{-4mm}
\section{Related Work}
\subsection{Feature Selection for Local Routing}
A classic feature for local routing comes from Greedy Forwarding \cite{finn1987routing}, in which the distance to the destination node is used to optimize forwarder selection. It has been proven that this feature achieves nearly optimal routing for certain network classes, including scale-free networks \cite{kleinberg2000navigation, papadopoulos2010greedy}. 
Notable other features for forwarder selection include Most Forward within Radius (MFR) \cite{takagi1984optimal}, Nearest with Forwarding Progress (NFP) \cite{hou1986transmission}, the minimum angle from neighbor $u$ to the line between origin $O$ and destination $D$, $\angle uOD$, (aka Compass Routing) \cite{kranakis1999compass}, and Random Progress Forwarding (RPF) \cite{nelson1984spatial}.  

Moreover, a recent study \cite{chen2023qf} shows that searching for shortest paths in uniform random graphs can be restricted to an elliptical search region with high probability. Its geographic routing protocol, {\it QF-Geo}, uses Node Stretch $\frac{\overline{O u}+\overline{u D}}{\overline{O D}}$, the stretch factor of node $u$ from the origin-destination-line $\overline{O D}$, as an input feature to determine whether a node's neighbor $u$ lies in the search region and to forward packets only within the predicted elliptical region.
We refer to this domain knowledge as {\em Elliptical Region for Bounded Search}.


\begin{figure}[t!]
    \centering
    \vspace{-4mm}
     \subfigure
     {
        \includegraphics[width=0.6\textwidth]{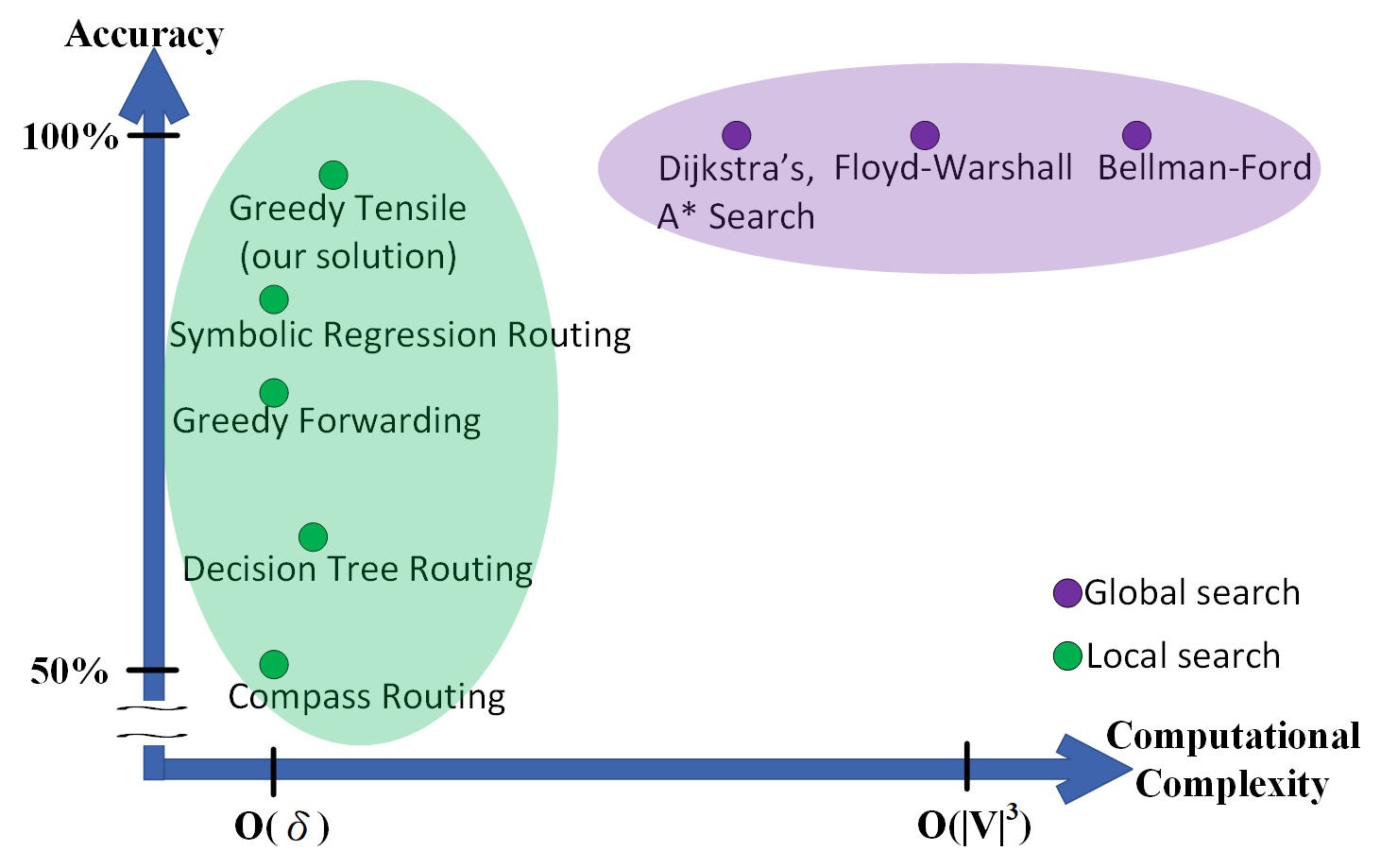}
    }
    \vspace{-4mm}
    \caption{A comparison of different (machine learned) routing policies for the per-hop best forwarder prediction for all-pair near shortest path problem (APNSP).}
    \label{routing_comp}
\end{figure}
\subsection{Local versus Global Routing Policies}

Figure~\ref{routing_comp} compares the accuracy and computational complexity\footnote{The computational complexity resulting from the global states is related to the network size $|V|$ and the number of edges $|E|$. Floyd–Warshall algorithm solves all-pairs (near) shortest paths with time complexity $O(|V|^{3})$.  Dijkstra's, A* search, and Bellman–Ford algorithms, which find single-source shortest paths, can be extended to solve APNSP by iterating $|V|$ sources and then incur the time complexity: $O(|V|^{2}\log{|V|})$, $O(|V|^{2}\log{|V|})$, $O(|V|^{2}|E|)$, respectively.} of \GreedyTensile with other routing policies based on local states or global states. 
With respect to the accuracy (in terms of near-shortest paths) for predicting the best forwarder, local routing policies yield diverse performances across machine-learned models or geographic routing policies. We explored symbolic regression
\cite{billard2002symbolic} routing and decision tree routing to select a subset of all candidate features\footnote{For a given node $u$ and its origin-destination node pair $(O, D)$, the feature set includes: distance to the destination $\overline{u D}$ , node stretch $\frac{\overline{O u}+ \overline{u D}}{\overline{O D}}$, node degree, and node angle $\angle uOD$.} that are significant to predict the optimal routing metric: both tended to choose node stretch as their primary feature, but did not yield a model whose accuracy across diverse network configurations (of sizes and densities) was ideal. We recall that extant geographic routing techniques that use a simple linear function over only one feature, such as Greedy Forwarding (using distance) and Compass Routing (using angle), do not achieve generalizability across various network topologies either. \GreedyTensile \!\!\!, which considers both distance-to-destination and node stretch as the input features, allows for searching a larger decision space and thus learns a model that is more sophisticated than a linear regression function and that achieves a near-optimal prediction.

\subsection{Generalizability of Machine Learned Routing}
Only recently has machine learning research started to address generalizability in routing contexts. For instance, generalizability to multiple graph layout distributions, using knowledge distillation, has been studied for a capacitated vehicle routing problem \cite{DBLP:conf/nips/Bi0WCCSC22}. Some explorations have considered local states: i.e., wireless network routing strategies using local states based on deep reinforcement learning \cite{manfredi2021relational,  manfredi2022learning} have been shown to generalize to other networks of up to 100 nodes, in the presence of diverse dynamics including node mobility, traffic pattern, congestion, and network connectivity. Deep learning has also been leveraged for selecting an edge set for a well-known heuristic, Lin-Kernighan-Helsgaun (LKH), to solve the Traveling Salesman Problem (TSP) \cite{DBLP:conf/nips/XinSCZ21}.  The learned model generalizes well for larger (albeit still modest) sized graphs and is useful for other network problems, including routing. Likewise, graph neural networks and learning for guided local search to select relevant edges have been shown to yield improved solutions to the TSP \cite{GNN-GLS-TSP}. In related work, deep reinforcement learning has been used to iteratively guide the selection of the next solution for routing problems based on neighborhood search \cite{wu2021learning}.

\section{Problem Formulation for Generalized Routing}
Consider the class $\mathbb{G}$ of all graphs $G = (V, E)$ whose nodes are uniformly randomly distributed over a 2-dimensional Euclidean geometric space. Each node $v \in V$ knows its global coordinates. For each pair of nodes $v, u \in V$, edge $(v, u) \in E$ holds if and only if the distance between $v$ and $u$ is at most the communication radius $R$, a user-defined constant. 

Let $\rho$ denote the network density, where network density is defined to be the average number of nodes per $R^2$ area, and $n$ the number of nodes in $V$. It follows that if all nodes in $V$ are distributed in a square area, its side must be of length $\sqrt{\frac{n \times R^2}{\rho}}$.

\subsection{All-Pairs Near-Shortest Path Problem (APNSP)}

The objective of APNSP routing problem is to locally compute for all node pairs of any graph $G \in \mathbb{G}$ their near-shortest path. Here, near-shortest path is defined as one whose length is within a user-specified factor ($\geq 1$) of the shortest path length.

Formally, let $d_e(O, D)$ denote the Euclidean distance between two endpoints $O$ and $D$, and $d_{sp}(O, D)$ denote the length of the shortest path between these endpoints. Further, let $\zeta(O, D)$ denote the path stretch of the endpoints, i.e., the ratio $\frac{d_{sp}(O, D)}{d_{e}(O, D)}$.

\noindent
\textbf{The APNSP Problem:} 
Learn a routing policy $\pi$ such that, for any graph $G = (V, E) \in \mathbb{G}$ and any origin-destination pair $(O, D)$ where $O, D \in V $,  $\pi(O, D, v) \! = \! u$ finds $v$'s next forwarder $u$ and in turn yields the routing path $p(O, D)$ with path length $d_{p}(O, D)$ that with high probability is a near-shortest path. In other words, $\pi$ optimizes the accuracy of $p(O, D)$ as follows:
\begin{align}
\max& ~~ Accuracy_{G, \pi} = \frac{\sum_{O, D \in V}{\eta(O, D)}}{|V|^2},
\\
s.t. & ~~
\eta(O, D) = \begin{cases} 1, \ \ i\!f \ \frac{d_{p}(O, D)}{d_{sp}(O, D)} \leq  \zeta(O, D) (1+\epsilon) \\
0, \ \ otherwise
\end{cases}
\label{APNSP_Accuracy}
\end{align} 

Note that the user-specified factor for APNSP is $\zeta(O, D) (1+\epsilon)$, where $\epsilon\geq 0$. The use of path stretch $\zeta(O, D)$ in Equation~\ref{APNSP_Accuracy} is to normalize the margin for shortest paths prediction (given $\epsilon$ is constant) across networks with different densities since the variance of path stretch becomes more significant as the network density $\rho$ decreases\footnote{The result in \cite{chen2023qf} shows that, in sparse graphs (e.g., $\rho$=1.4, 2), the path stretch can vary in $[1.0, 10.0]$ since some of the shortest paths are prone to be found around the holes. However, the path stretch in dense graphs (e.g., $\rho$=4, 5) is likely to vary only within the range of $[1.0, 3.5]$. Thus, we exploit $\zeta(O, D)$ to mitigate the gap between sparse and dense networks for APNSP prediction.}.

\subsection{MDP Formulation for the APNSP Problem}
\begin{figure}[thb!]
    \centering
     \vspace{-6mm}
     \subfigure
     {
        \includegraphics[width=0.7\textwidth]{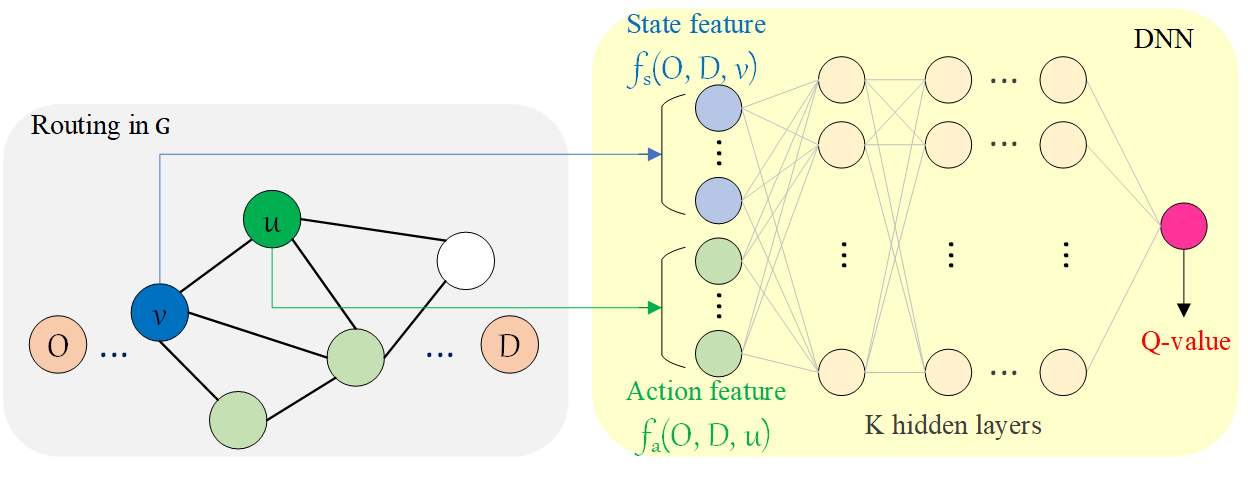}
    }
    \vspace{-4mm}
    \caption{Schema for solution using DNN to predict $Q$-value for selecting the routing forwarder.}
    \label{DNN}
\end{figure}
To solve the APNSP problem, we first formulate it as a Markov decision process (MDP), which learns to choose actions that maximize the expected future reward. In general, an MDP consists of a set of states $S$, a set of actions  $A$ for any state $s\in S$, an instantaneous reward $r(s, a)$, indicating the immediate reward for taking action $a\in A$ at state $s$, and a state transition function $P(s'|s, a)$ that characterizes the probability that the system transits to the next state $s'\in S$ after taking action $a$ at the current state $s\in S$. To simply the routing behavior in the problem, the state transition is assumed to be deterministic. Specifically, each state $s$ represents the features of a node $v$ holding a packet associated with an origin-destination pair $(O, D)$, and an action $a\in A=nbr(v)$ indicates the routing behavior to forward the packet from node $v$ to one of its neighbors $u \in nbr(v)$, where $nbr(v)$ denotes the set of $v$'s one-hop neighbors. Given the current state $s$ and an action $a\in nbr(v)$ which selects one neighbor as the next forwarder, the next state $s'$ is determined as the features of the selected neighbor $u$ such that the probability $P(s'|s, a)$  is always one. The tuple $(s, a, r, s')$ is observed whenever a packet is forwarded. 

\vspace*{1.5mm}
\noindent
{\bf Design of $Q$:}
In general, the $Q$-value is defined to specify the cumulative future reward from state $s$ and taking action $a$: 
\vspace*{-2mm}
\begin{align*}
    Q(s_t=s, a_t=a) = \sum_{i=t}^L \gamma^{i-t} r(s_i, a_i)
    \vspace*{-2mm}
\end{align*}
where $\gamma, 0 \! \leq \! \gamma \! \leq \! 1,$ is the discount factor. When $\gamma=0$, the instantaneous reward is considered exclusively, whereas the future rewards are treated as equally important as the instantaneous reward in the $Q$-value if  $\gamma=1$. 

To design the reward function $r$, we exploit the domain knowledge of {\em Elliptical Region for Bounded Search} to prioritize forwarding that yields a routing path within an ellipse. In the APNSP problem, the user-specified factor, $\zeta(O, D) (1+\epsilon)$, implies the size of the ellipse for search. A routing path $p$ between $O$ and $D$ with the length $d_{p}(O, D)$, not larger than $d_{sp}(O, D)\zeta(O, D) (1+\epsilon)$, is considered a near-optimal path. For any action that will make the routing path stretch outside the elliptical region, a negative penalty is added to the reward $r$. Note that, in supervised learning, the path stretch $\zeta(O, D)$ for each source-destination pair is known for calculating the optimal $Q$ values. In Reinforcement Learning (RL), the probabilistic bound of the path stretch (i.e., the size of the ellipse for search) can be modeled as a function of network density and the Euclidean distance between the two endpoints \cite{chen2023qf}. 

Therefore, we define the instantaneous reward as follows:
\begin{align*}
    r(s, a) =  \begin{cases} -d_{e}(s, s'), \ \ i\!f \ \frac{d_{e}(s, s')+d_{sp}(s', D)}{d_{sp}(s, D)} \leq  \zeta(O, D) (1+\epsilon) \\
    -d_{e}(s, s') - C\Delta(s, s', O, D), \ \ otherwise
    \end{cases},
\end{align*}
where $\Delta(s, s', O, D) \geq 0$ is the penalty to the reward and $C>0$ is a constant. $\Delta(s, s', O, D) = (d_{e}(s, s')+d_{sp}(s', D)) - d_{sp}(s, D)\zeta(O, D) (1+\epsilon)$ calculates the part that exceeds the margin. We set $\gamma=1$, and, therefore, the optimal $Q$-value $Q^*(s, a)$ is equal to the cumulative negative length of the shortest path from $s$ to the destination through $s'$, if the path is near-optimal.

To solve the APNSP problem, we seek to learn the optimal $Q$-value through a data-driven approach with DNNs. As depicted in Figure~\ref{DNN}, each state $s$ and action $a$ will be embedded into a set of input features denoted by $f_{s}(s)$ and $f_{a}(a)$\footnote{Since $O$ and $D$ will remain the same in all forwarder prediction associated with one $(O, D)$ pair, for notational convenience, we will omit $O$ and $D$ in the DNN input and use $f_{s}(v)$ and $f_{a}(u)$ to represent the state and action features.}, respectively. A DNN will be learned to approximate the optimal $Q$-value given the input features, based on which a near-shortest path routing policy can be obtained by taking actions with largest $Q$-values. We begin with theory that guides the learnability of a generalizable DNN for APNSP.

\section{Explainable Learnability and Generalizability of Routing Policy}
\label{section_theory}

In this section, we identify two sufficient conditions, {\bf Pointwise Monotonicity} and {\bf RankPres} that, respectively, imply learnability and generalizability of an optimal routing policy. We also identify domain knowledge on the satisfiability of these two conditions in the context of the APNSP routing problem. Together, these explain the feasibility of learning a routing policy from limited data on a few nodes in a single graph that suffices for all graphs $G \in \mathbb{G}$. (For reasons of space, all proofs of the theory in this section are relegated to Appendix~\ref{appendix_proof}.)

The basic concept in the APNSP context is of a ranking metric function for each node $v \! \in \! V$ with respect to each $u \! \in \! nbr(v)$. Let $f_{s} \! : \! V \! \rightarrow \! \mathbb{R}^{I}$ be a map of $v \in V$ to its state features (of cardinality $I$) and let $f_{a} \! : \! V \! \rightarrow \! \mathbb{R}^{J}$ be a map of $u \! \in \! nbr(v)$ to its action features (of cardinality $J$). We define a ranking metric $m \! : f_{s} \times f_{a} \! \rightarrow \! \mathbb{R}$
to be a {\em linear} or {\em non-linear} function over the input features associated with a node $v$ and its neighbor $u$.



\subsection{Learnability of Ranking Metric Function $m$}


\begin{definition}[Pointwise Vector Ordering $\leq$]
A vector ordering $\leq$ is pointwise if for any two vectors, $A = (a_1, a_2, ..., a_n) \in \mathbb{R}^{n}$ and $B = (b_1, b_2, ..., b_n) \in \mathbb{R}^{n}$, $A \leq B$ iff $(\forall k: 1 \leq k \leq n : a_k \ \leq b_k$).
\end{definition}


\begin{definition}[Pointwise Monotonicity of $m$ wrt $\leq$]
For any node $v$ in $V$ and ranking metric $m$, $m$ is {\bf Pointwise Monotonic} with respect to a pointwise vector ordering $\leq$ iff 
\begin {quote}
For any ordering $\langle u_1, ..., u_d \rangle$ of all nodes in $nbr(v)$, if $ (f_{s}(v), f_{a}(u_1)) \leq ... \leq (f_{s}(v), f_{a}(u_d))$ then $m(f_{s}(v), f_{a}(u_1)) \leq ... \leq m(f_{s}(v), f_{a}(u_d)$

\end{quote}
\end{definition}

\begin{theorem}
[Learnability]
Assume the metric $m$ is Pointwise Monotonic with respect to $\leq$ for a node $v$ in $V$. Then there exists a DNN $H$, $H\!: \mathbb{R}^{n} \! \rightarrow \! \mathbb{R}$, with training samples $\langle X_{v}, Y_{v} \rangle$ that learns $m$ for $v$.
\label{Learnability}
\end{theorem}

\subsection{Generalizability of Ranking Metric Function $m$}

We next define property {\bf RankPres} that relates the local ranking metric $m$ and the corresponding $Q$-values set $Y_{v}$, namely, the global ranking metric. RankPres  suffices to apply $m$ for ranking the neighbors $u \in nbr(v)$ using local node states to achieve the same ranking using $Q$-values, which allows routing to be performed locally as well as is achieved by using $Q$.

\begin{definition} [RankPres] A metric function satisfies {\em RankPres} for a node $v$ in $V$ iff 
\begin {quote}
For any ordering $\langle u_1, ..., u_d \rangle$ of all nodes in $nbr(v)$, if $\langle m(f_{s}(v), f_{a}(u_1)), \, ... \, , \\ \, m(f_{s}(v), f_{a}(u_d) \rangle$ is monotonically non-decreasing, then $\langle Q(v, u_1), ..., Q(v, u_d)\rangle$ is monotonically non-decreasing. 
\end{quote}
\end{definition}

Next, we lift the sufficient condition to provide a general basis for ranking the neighbors of all nodes in a graph to predict the best forwarder, by training with only the samples derived from one (or a few) of its nodes.  Subsequently, we further lift the sufficient condition for similarly ranking the neighbors of all graphs in $\mathbb{G}$.

For notational convenience, given an ordering $\langle u_1, ..., u_d \rangle$ of all nodes in $nbr(v)$, let $X_{v} = \{\langle f_{s}(v),f_{a}(u_1) \rangle, ..., \langle f_{s}(v),f_{a}(u_d) \rangle\}$, denote the set of vectors for each corresponding neighbor $u_k \! \in \! nbr(v), 1 \! \leq \! k \! \leq d$. Also, let $Y_{v} = \{Q(v, u_1), ..., Q(v,u_d)\}$ denote the corresponding set of $Q$-values.

\begin{lemma}
[Cross-Node Generalizability] 
\label{cross_node_generalizability}
For any graph $G$, $G \! = \! (V,E)$, if ranking metric $m(f_{s}(v), f_{a}(u))$ satisfies RankPres for all $v \! \in \! V$ and is learnable with a DNN $H$ trained with only a subset of training samples $\langle X_{V'}, Y_{V'} \rangle$, where $V' \subseteq V$, $X_{V'} \! = \! \bigcup_{v \in V' }{X_{v}}$, and $Y_{V'} \! = \!  \bigcup_{v \in V'}{Y_{v}}$, then $H$ approximates an optimal ranking policy for all $v \! \in \! V$.
\end{lemma}


Note that if the $Q (v, u)$ value corresponds to the optimal (shortest) path $Q$-value for each $(v, u)$ pair, then $m$ indicated by Lemma~\ref{cross_node_generalizability} achieves an optimal routing policy for all nodes in $V$.  
Note also that if  $m$ satisfies RankPres not for all nodes but for almost all nodes, a policy learned from samples from one or more nodes $v$ that satisfy RankPres may not achieve optimal routing for all nodes. Nevertheless, if the relative measure of the number of nodes that do not satisfy RankPres to the number of nodes that do satisfy RankPres is small then, with high probability, the policy achieves near-optimal routing.

\begin{lemma}
[Cross-Graph Generalizability]
If metric $m(f_{s}(v), f_{a}(u))$ satisfies RankPres for the nodes in all graphs $G \! \in \! \mathbb{G}$ and is learnable with a DNN $H$ trained with samples from one or more nodes in one or more chosen seed graph(s) $G^{*} \! \in \! \mathbb{G}$, then $H$ approximates an optimal ranking policy for all graphs $G$.
\label{cross_graph_generalizability}
\end{lemma} 

Again, if Lemma~\ref{cross_graph_generalizability} is considered in the context of $Q$-values corresponding to optimal shortest paths, 
the learned routing policy $H$ generalizes to achieving optimal routing over all graphs $G \! \in \! \mathbb{G}$. And if we relax the requirement that {\bf RankPres} holds for all nodes of all graphs in $\mathbb{G}$ to only require that for almost all graphs $G \! \in \! \mathbb{G}$, there is a high similarity between the ranking orders of neighbors using $m$ and the ones using optimal $Q$-value over all nodes $v$, then with high probability the policy achieves near-optimal routing.



\subsection{Learnability and Generalizability for APNSP}


\begin{theorem}[Metric Function for Optimal Policy]
If a metric function $m$ satisfies RankPres and Pointwise Monotonicity, it yields an optimal generalizable and learnable ranking policy.
\end{theorem}


We now instantiate the theory for provable learnability and generalizability for APNSP.
\begin{proposition}
For APNSP, there exists a ranking metric $m_1(f_{s}(v), f_{a}(u))$ of the form $w_{1} .  d(v, D) + w_{2} . d(u, D)$ based on the {\bf distance-to-destination} input feature that satisfies the property of RankPres and Pointwise Monotonicity for almost all nodes in almost all graphs $G$. \\ \ \\
Also, there exists a ranking metric $m_2(f_{s}(v), f_{a}(u))$ of the form $w_{1} . d(v, D) + w_{2} . ns(O, D, v) + w_{3} . d(u, D) + w_{4} . ns(O, D, u)$ based on both {\bf distance-to-\\destination} and {\bf node stretch} input features that satisfies the property of RankPres and Pointwise Monotonicity for almost all nodes in almost all graphs $G$.
\label{proposition_1}
\end{proposition}

In Appendix~\ref{ranking_similarity}, we present empirical validation of Proposition~\ref{proposition_1} for graphs in Euclidean space.
We respectively choose two linear functions, $m_1$ and $m_2$, that satisfy Pointwise Monotonicity. And then RankPres is quantified in terms of Ranking Similarity; high ranking similarity for both $m_1$ and $m_2$ implies that RankPres holds with high probability. 
It follows that an efficient generalizable policy for APNSP is learnable for each of the two chosen input feature sets.

\section{Single Graph Learning}
\subsection{Design of Input Features}

To learn a routing protocol that generalizes across graphs with different scales, densities, and topologies, the input features of the DNN should be designed to be independent of global network configurations, including the identity of nodes and of packets. 
Recall that each node knows its own coordinates and the coordinates of the origin and the destination. 

For feature selection, we first apply the decision tree and symbolic regression to filter the crucial features from a set of candidate features widely used in geographic routing protocols: {distance from $v$ to destination $D$, perpendicular distance from node $v$ to the origin-destination-line $\overleftrightarrow{O D}$, angle between $\overline{v O}$ and $\overline{O D}$, and node stretch relative to $(O, D)$ pair}. Note that symbolic regression finds a mathematical function (using a combination of an operator set $\{+, -, *, /, cos, sin, log, min, max, ...\}$) that best describes the relationship between the input features and the output on a given dataset. In addition to the distance from $v$ to destination $D$ that is used to achieve near-optimal routing in scale-free graphs, node stretch is found to be the dominant feature for predicting the $Q$-values using both decision tree and symbolic regression. These results motivate us to adopt these two features as the node states for the best forwarder prediction for APNSP.
Accordingly, we design the input features, including state and action features, as follows:
\vspace*{-1.5mm}
\begin{itemize}
\item State feature, $f_{s}(v)$. 
For a packet with its specified origin $O$ and destination $D$ at node $v$, the state features are the vectors with the elements below.
    \begin{enumerate}
    \item \textbf{Distance to destination, $d(v, D)$:} the distance between  $v$ and $D$.
    \item \textbf{Node Stretch, $ns(O, D, v) = \frac{d(O, v)+d(v, D)}{d(O, D)}$:} the stretch of the indirect distance between $O$ and $D$ that is via $v$ with respect to the direct distance between $O$ and $D$. 
    \end{enumerate}

\item Action feature, $f_{a}(a) = f_{s}(u)$. The feature for the action that forwards a packet from $v$ to $u \in nbr(v)$, $f_{a}(a)$ is chosen to be the same as the state feature of $u$, $f_{s}(u)$. 
\end{itemize}

Henceforth, we consider learning with two different combinations of input features, one with only $d(v, D), d(u, D)$ and the other with both $d(v, D), d(u, D)$ and $ns(O, D, v), ns(O, D, u)$.

\subsection{Selection of Seed Graph and Graph Subsamples}
To achieve both cross-graph generalizability and cross-node generalizability, we develop a knowledge-guided mechanism with the following two selection components: (1) seed graph selection and (2) graph subsample selection.

{\bf Seed graph selection.} 
The choice of seed graph depends primarily on the analysis of cross-node generalizability (Lemma~\ref{cross_node_generalizability})
across a sufficient set of uniform random graphs with diverse sizes and densities/average node degrees. 
In Figure~\ref{d_sf_graph_similarity_euclidean}, we empirically show that, with the use of distance-to-destination and node stretch, a good seed graph (with high $SIM_{G}(m, Q^{*})$) is likely to exist in a set of graphs with modest size (e.g., 50) and high density (e.g., 5) in the Euclidean space. 

There may be applications where analysis of large (or full) graphs is not always possible. In such situations, given a graph $G$, an alternative choice of seed graph can be from a small subgraph $G'=(V', E'), V' \subset V, E' \subset E$ with relatively high cross-node generalizability. Note that, in Lemma~\ref{cross_node_generalizability}, for a graph $G=(v,E)$ satisfying the {\bf RankPres} property for all $v \in V$, {\bf RankPres} still holds for $v' \in V'$ in a subgraph $G'=(V', E')$ of $G$. This is because the learnable function $m$ still preserves the optimal routing policy for all nodes in a subset of $nbr(v)$.

{\bf Graph subsamples selection.} 
We provide the following scheme of subsample selection for a given graph $G=(V,E)$ to choose a set of $\phi$ nodes for generating $\phi\delta$ training samples, where $\delta$ is the average node degree (i.e., number of neighbors).

\begin{enumerate}
\item Select an origin and destination pair $(O, D), O, D \in V$.

\item Select $\phi$ nodes, $v_{1}, ..., v_{\phi} \in V \setminus {D}$.

\item For each chosen node $v_{1}, ..., v_{\phi}$, respectively, collect the subsamples $\langle X, Y \rangle$, where \\ $X \! = \! \bigcup_{v \in \{v_{1}, ..., v_{\phi}\} \wedge u \in nbr(v)}{\{\langle f_{s}(v),f_{a}(u)\rangle \}}$ and \\ $Y \! = \!  \bigcup_{v \in \{v_{1}, ..., v_{\phi}\} \wedge u \in nbr(v)}{\{Q^{*}(v, u)\}}$.
\end{enumerate}

\subsection{Supervised Learning for APNSP with Optimal $Q$-values}

Given the dataset $\langle X, Y\rangle$ collected by using the subsampling policy for the seed graph, we train a DNN based on supervised learning to capture the optimal ranking policy. Specifically, suppose the DNN $H$ is parameterized by $\theta$. We seek to minimize the following loss function:
\begin{align}\label{supervised}
    \min_{\theta}~~\sum_{\langle X, Y\rangle} \|H_{\theta}(f_{s}(v),f_{a}(u)) - Q^*(v,u)\|^2.
\end{align}
Note that we assume that the optimal $Q$-values are known for the seed graph in the supervised learning above, which can be obtained based on the shortest path routing policies of the seed graph. By leveraging these optimal $Q$-values and supervised learning on the seed graph, a generalized routing policy is learned for APNSP routing over almost all uniform random graphs, as we validate later in the experiments.

\vspace*{-2mm}
\subsection{Reinforcement Learning for APNSP}
\vspace*{-2mm}
For the case where the optimal $Q$-values of graphs are unknown, we  develop an approach for solving the APNSP problem based on RL. Using the same input features and the same seed graph selection procedure, the RL algorithm continuously improves the quality of $Q$-value estimations by interacting with the seed graph.  

In contrast to the supervised learning algorithm, where we collect only a single copy of data samples from a set of chosen (shortest path) nodes once before training, in RL new training data samples from nodes in a shortest path, predicted by most recent training episode (i.e., based on the current $Q$-value estimation), are collected at the beginning of each training episode. Remarkably, the generalizability of the resulting RL routing policy across almost all graphs in $U$ is preserved.
The details of the RL algorithm, named as RL-APNSP-ALGO, are shown in Algorithm~\ref{RL_ALGO} in Appendix~\ref{appendix_RL_algo}.

\section{Routing Policy Performance for Scalability and Zero-shot Generalization}
In this section, we discuss implementation of our machine learned routing policies and evaluate their performance in predicting all-pair near-shortest paths for graphs across different sizes and densities in Python3. We use PyTorch 2.6.0 \cite{pytorch} on the CUDA 12.4 compute platform to implement DNNs as shown in Figure~\ref{DNN}. Table~\ref{SimulationParams} in Appendix~\ref{appendix_results} shows our simulation parameters for training and testing the routing policies.

\vspace*{-2mm}
\subsection{Comparative Evaluation of Routing Policies}
\vspace*{-2mm}

We compare the performance of the different versions of \GreedyTensile policies obtained using the following approaches:

\begin{enumerate}
\item {\bf GreedyTensile-S}:  Supervised learning from appropriately chosen seed graph $G^{*}$ using graph subsamples selection from $\phi$=3.

\item {\bf GreedyTensile-RL}: RL from appropriately chosen seed graph $G^{*}$ using graph subsamples selection from $\phi$=3.

\item {\bf GF}: Greedy forwarding that forwards packets to the one-hop neighbor with the minimum distance to the destination. 

\item {\bf SR-NS}: Symbolic Regression routing that learns a function, $ns(O, D, u)+0.64$, to forward packets to the one-hop neighbor with the minimum node stretch. 
\end{enumerate}

Note that for a given set of input features, both supervised learning and reinforcement learning schemes use the same DNN configuration to learn the routing policies. By using the subsampling mechanism, not only the sample complexity but also the training time will be significantly reduced in GreedyTensile-S and GreedyTensile-RL since they only require the data samples derived from $\phi$=3 nodes rather than all nodes in a seed graph.

\subsection{Zero-shot Generalization over Diverse Graphs}
\begin{figure*}[hbt!]
    \centering
    \vspace{-8mm}
    \subfigure
    {
    \includegraphics[width=0.6\textwidth]{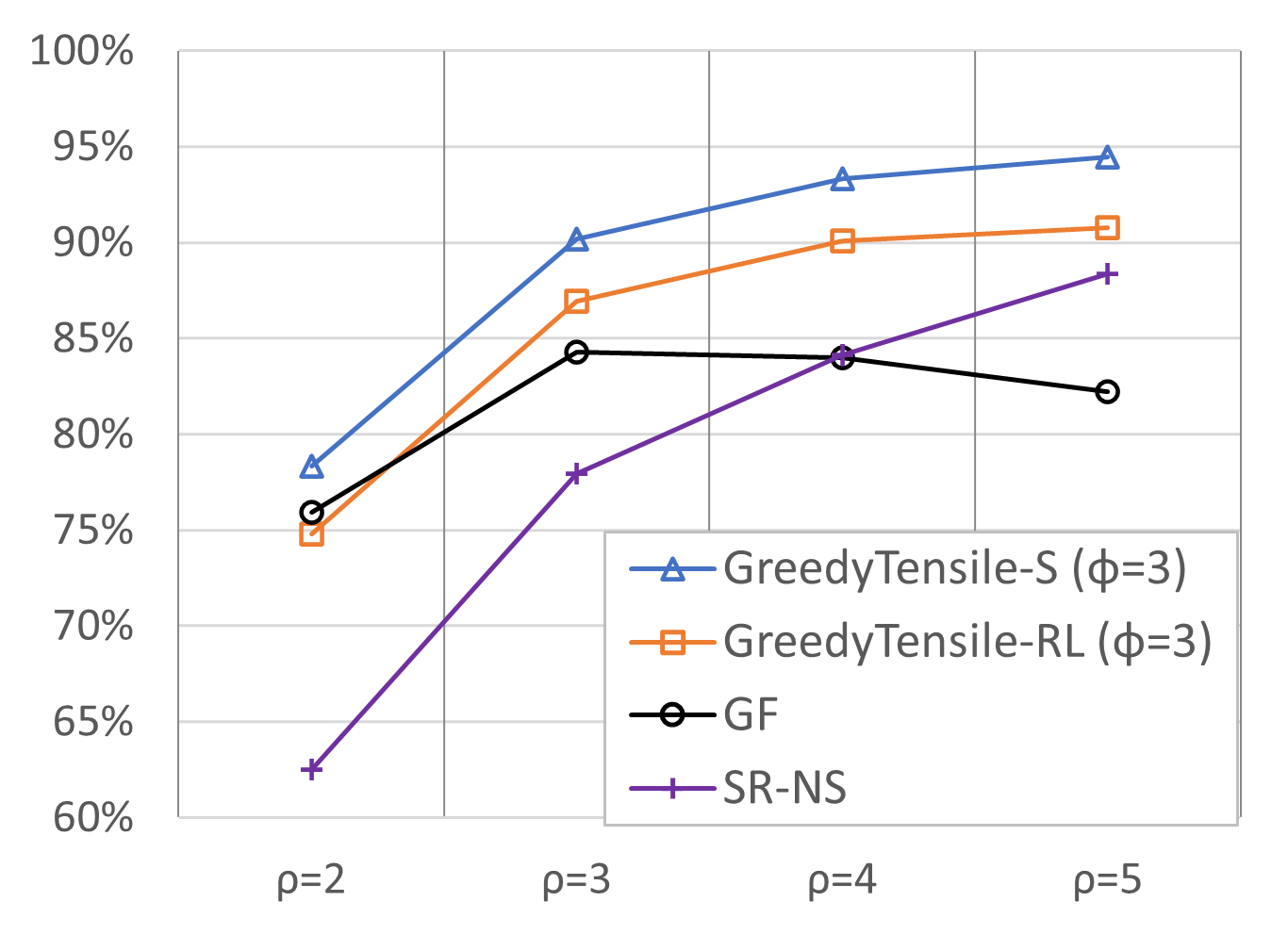}
        \label{testing_euclidean}
    }
    \vspace{-4mm}
    \caption{Average APNSP prediction accuracy across graph sizes with various $\rho$ in Euclidean space for \GreedyTensile policies.}
    \label{Testing_Performance_ed_sf}
\end{figure*}

To evaluate the scalability and generalizability of the policies, we directly (i.e., without any adaptation) test the policies learned from the seed graph $G^*$ on new uniform random graphs with different combinations of $(N_{test}, \rho_{test})$ in Euclidean spaces. We select 80 random graphs for each pair and calculate the average prediction accuracy over these $80{N_{test}}^2$ shortest paths. 

For the DNNs with input  $d(v, D)$ and $d(u ,D)$, the tests confirm that the performance of all the learned policies exactly match the prediction accuracy of GF. 

For the \GreedyTensile DNNs with input $\langle d(v, D), ns(O, D, v), d(u, D), \\ ns(O, D, u)\rangle$, we plot in Figure~\ref{Testing_Performance_ed_sf} the respective average prediction accuracies across graph sizes in \{27, 64, 125, 216\} with density in \{2, 3, 4, 5\}. 

The Supervised approach (GreedyTensile-S)  achieves the best performance among all the approaches. In particular, compared to GF,  GreedyTensile-S improves the accuracy by up to 12. 22\% over GF, while the RL approach \\ (GreedyTensile-RL) shows at least comparable performance in low-density graphs ($\rho=2$) and achieves an improvement of up to 8.55\% in graphs with $\rho \geq 3$. The performance gap between DNNs and GF increases as the network density increases to a high level (e.g., $\rho=5$), where GF was believed to work close to optimal routing. SR-NS, merely generalizing to a specific network configuration, yields an accuracy gap of up to 12. 42\% over GF in low-density graphs ($\rho \leq 2$) and starts to outperform GF in high-density graphs ($\rho \geq 4$). The performance of SR-NS increases to come close to the performance of the GreedyTensile-RL policy as the density increases to $\rho=5$.

The results imply that, comparing the features of distance-to-destination and node stretch, routing using only distance-to-destination (i.e., GF) or routing using only node stretch (i.e., SR-NS) tend to generalize only in sparse or dense wireless networks, respectively. Our DNNs take advantage of these two features and produce more sophisticated policies based on the current node and $(O, D)$ information by searching for a larger input space, and then show comprehensive generaliability across diverse network configurations. 

We further compare GreedyTensile DNNs with datasets derived from chosen nodes versus all nodes to evaluate the effect of graph subsampling in Appendix~\ref{ablation_study}. The results show that exploiting domain knowledge in sample selection yields DNNs with not only higher sample efficiency but higher performance in APNSP prediction accuracy.


\vspace*{-2mm}
\section{Discussion}
\vspace*{-2mm}
{\bf Stabilization}. The learned policy is self-stabilizing in a straightforward manner: the routing decision using \GreedyTensile depends only on the local network states, $\langle f_{s}(v), f_{a}(u) \rangle$, $u \in nbr(v)$. \GreedyTensile thus recovers from arbitrarily inconsistent or stale states (e.g., due to topology changes or crashed nodes) by locally retrieving the coordinates of the current neighbors. 

\noindent
{\bf Handling of inaccurate routes}. For handling of route failures, i.e., when the route leads to a dead-end, the policy can be extended by taking advantage of the knowledge that searching within the elliptical region will, with high probability, yield a path between the source and destination. As is typical, depth-first search (DFS) can be used to traverse all nodes within the ellipse.  Let $G'=(V',E') \subseteq G$ be the subgraph including all nodes $v' \in V'$ and edges $e' \in E'$ within a given elliptical region. The time complexity of rediscovering a routing path using DFS is $O(|V'|+|E'|)$. 

\noindent
{\bf Generalizable routing for random geometric Hyperbolic space graphs}. 
We extend our experimental findings to show the zero-shot generalizability of \GreedyTensile for the APNSP problem in hyperbolic metric space. This space is known to represent the scale topologies of the Internet and social networks, where node degree follows a power-law distribution \cite{boguna2010sustaining, verbeek2014metric}. As with the Euclidean space, \GreedyTensile generalizes to all random hyperbolic graphs with different average node degrees. To the best of our knowledge, we are the first to provide a routing policy that outperforms GF in almost all random graphs; recall that GF has been shown to find almost optimal shortest path in scale-free topologies \cite{papadopoulos2010greedy}. The detailed results are shown in Appendix~\ref{appendix_results}.

\noindent
{\bf Symbolic interpretation of {\it Greedy Tensile}.}
We symbolically interpret the learned DNN with input of $d(v,D)$, $d(u,d)$, $ns(O,D,v)$, and $ns(O,D,v)$ to show that it can be approximated by a low-complexity two-linear action policy, which is visualized and elaborated upon in Appendix~\ref{interpretability}.  One action uses both distance-to-destination and node stretch features to select the next forwarder for those neighbors providing forward progress, and the other considers only distance-to-destination features in the absence of forward progress.

\vspace*{-2mm}
\section{Conclusions and Future Work}
\vspace*{-2mm}

We have shown that guiding machine learning with domain knowledge can lead (somewhat surprisingly) to the rediscovery of a well-known routing policy, in addition to discovering a new routing policy that performs well in terms of complexity, scalability, and generalizability. The theory we have presented in the paper is readily extended to other classes of graphs (such as non uniform cluster distributions) and MDP actions that span multiple neighbors. Thus, albeit our illustration intentionally uses relatively familiar input features and local routing architectures, richer domain theory will be useful to guide machine learning of novel routing algorithms. Moreover, the routing policies are likely to be competitive for richer classes of graphs than the class of uniform random graphs on which we have focused our validation.

While samples from nodes of a single seed graph suffice for generalizable learning, in practice, learning from multiple seed graphs may be of interest.  For instance, if an ideal seed graph is not known a priori, online learning from better or multiple candidate seed graphs as they are encountered may be of interest for some applications. Along these lines, we recall that the set of ideal (and near-ideal) seed graphs is relatively large in the problem we considered. One way to relax the knowledge of ideal seed graphs is to leverage online meta-learning, for learning a good model initialization and continuing to improve the initialization based on better seed graphs as they are encountered. Towards this end, we have also been studying the merits of efficiently fine tuning the model for the target graph as an alternative to zero-shot generalization.

%
%
%
\newpage
\bibliographystyle{splncs04}
\bibliography{mybibliography}

\appendix
\section{Proofs for Learnability and Generalizability}
\label{appendix_proof}
In this section, we provide the proof of the theorems of generalizability and learnability stated in Section~\ref{section_theory}.

\begin{enumerate}
\item Proof of Theorem~\ref{Learnability}.
\begin{proof}
The theorem follows directly from a construction \cite{sill1997monotonic} that demonstrates the learnability of neural networks when there exists a monotonic mapping, $H$, from the input vector to the output value. 
\end{proof}

We note that Pointwise Monotonicity of $m$ always holds for all $v \in V$ when $J=1$. Namely, the input dimension of the action features $f_a(u)$ is $1$, and state features $f_{s}(v)$ will stay unchanges in $X_{v}$. There is always an order of neighbors for all $v\in V$ that satisfies Pointwise Monotonicity of $m$. Any node $v \in V$ with degree greater than 1 suffices to learn $H$ that preserves this property.
However, when $J>1$, there may exist a node $v' \in V$ (with its neighbor $u' \in nbr(v')$), which does not satisfy the property Pointwise Monotonicity for $m$. In order to learn $H$ which achieves the same ranking as $m$ does, the training samples $\langle X_{v}, Y_{v} \rangle$ derived from $v$ should be appropriately chosen to ensure the learnability of $m$. \\

\item Proof of Lemma~\ref{cross_node_generalizability}.
\begin{proof}

Our objective is to learn from a sample subset $\langle X_{V'}, Y_{V'} \rangle$ a DNN, $H: f_{s} \times f_{a} \rightarrow \mathbb{R}$, that satisfies RankPres property for all nodes in a graph. 
Since there exists $m()$ for which RankPres holds for all nodes, then from Theorem~\ref{Learnability}, a DNN $H: f_{s} \times f_{a} \rightarrow \mathbb{R}$ can be learned from a subset of nodes $V'$, $V' \subseteq V$ and $|V'| \geq 1$. In fact, it suffices to learn $H$ with the training samples derived from a single $v \in V_{m} \subseteq V $ with the highest degree (degree($v$) $>$ 1), where $V_{m}$ is the set of nodes satisfying Pointwise Monotonicity of $m$.
It remains to show that $H$ satisfies RankPres on nodes $v'' \in V \setminus V'$.
Without loss of generality, let us assume that $Q(v'', u_1'')$, $Q(v'', u_2'')$, \ldots are in monotonic non-decreasing order. Since $H$ satisfies Pointwise Monotonicity, for this neighbor ordering, by definition, the outputs of $H$ will also be in the same order as that of $Q$.  



\end{proof}

\item Proof of Lemma~\ref{cross_graph_generalizability}.
\begin{proof}
If there exists $m()$ that satisfies RankPres property for the nodes in all graphs, then a DNN $H: f_{s} \times f_{a} \rightarrow \mathbb{R}$ can be learned by using training samples from one or more nodes in one or more chosen seed graph(s) $G^*$. It is sufficient to learn $H$ with the training samples derived at least from one node with the highest degree in the seed graph $G^*$. $H$ achieves the optimal ranking policy.
\end{proof}

\end{enumerate}

\section{Ranking Similarity between Local and Global Metrics}
\label{ranking_similarity}

Based on the theory of Section~\ref{section_theory}, we now investigate the existence of a ranking metric $m$ for the first input feature, as well as the pair of input features, and demonstrate that in both cases the ranking similarity is close to 1 across nodes in almost all graphs $G$, regardless of their size and density.  The results thus empirically corroborate the feasibility of routing policy generalization from single graph learning, and also guide the selection of seed graphs and training samples.

Let $SIM_{v}(m, Q^{*}) \in [0,1]$ denote the {\em ranking similarity} between the ranking metric $m$ and the optimal $Q$-values ($Q^{*}$) at node $v$.  Moreover, let $SIM_{G}(m, Q^{*}) \in [0,1]$ denote the average ranking similarity between $m$ and $Q^{*}$ across all nodes $v \in V$ in a given graph $G=(V,E)$, i.e., $SIM_{G}(m, Q^{*}) = \frac{\sum_{v \in V}{SIM_{v}(m, Q^{*})}}{|V|}$.

Conceptually, $SIM_{v}(m, Q^{*})$ should tend to 1 as the order of neighbors in the sorted ascending order of $m(f_{s}(v), f_{a}(u)), u \in nbr(v)$, comes closer to matching the order of the neighbors in the sorted ascending order of $Q^{*}(v, u)$. More specifically, we adopt Discounted Cumulative Gain (DCG) \cite{jarvelin2002cumulated} to formally define the ranking similarity. The idea of DCG is to evaluate the similarity between two ranking sequences by calculating the sum of graded relevance of all elements in a sequence. First, a sorted sequence of $Q^{*}(v, u)$ with length $L = |nbr(v)|$, $u \! \in \!  nbr(v)$, is constructed as the ideal ranking, denoted by $A$. For each position $i$ in $A$, we assign a graded relevance $rel_{A}[i]$ by following the rule: $rel_{A}[i] = (L - i)^2, i = 0 ... L-1$\footnote{The assignment of graded relevance could be any way to decrease the value from left to right positions. Here we use squared value to assign dominant weights to the positions close to leftmost.}. The value of DCG accumulated at a particular rank position $\tau$ is defined as:
\begin{equation*}
\begin{array}{l}
DCG_{\tau} = \sum_{i=1}^{\tau}{\frac{rel[i]}{\log_{2}(i+1)}}.
\label{DCG}
\end{array}
\end{equation*} 
For example, let $A = [4, 1, 3, 2, 5]$. The corresponding $rel_{A}[i]$ and $\frac{rel_{A}[i]}{\log_{2}(i+1)}$ values are shown in Table~\ref{ideal_DCG_table}. Then $A$'s $DCG_{3} = 25+10.095+4.5 = 39.595$.

\begin{table}[htb!]
\caption{An example of DCG calculation for an ideal ranking $A=[4, 1, 3, 2, 5]$.}
\begin{center}
\begin{tabular}{|c|c|c|c|c|}
\hline
\textbf{$i$} & \textbf{$A[i]$} & \textbf{$rel_{A}[i]$} & \textbf{$\log_{2}(i+1)$} & \textbf{$\frac{rel_{A}[i]}{\log_{2}(i+1)}$} \\
\hline
1 & 4 & 25 & 1 & 25\\
2 & 1 & 16 & 1.585 & 10.095\\
3 & 3 & 9 & 2 & 4.5\\
4 & 2 & 4 & 2.322& 1.723\\
5 & 5 & 1 & 2.807 & 0.387\\
\hline
\end{tabular}
\label{ideal_DCG_table}
\end{center}
\end{table}
\begin{table}[htb!]
\caption{An example of DCG calculation for an estimated ranking $B = [1, 2, 4, 5, 6]$.}
\begin{center}
\begin{tabular}{|c|c|c|c|c|}
\hline
\textbf{$j$} & \textbf{$B[j]$} & \textbf{$rel_{B}[j]$} & \textbf{$\log_{2}(j+1)$} & \textbf{$\frac{rel_{B}[j]}{\log_{2}(j+1)}$} \\
\hline
1 & 1 & 16 & 1 & 16\\
2 & 2 & 4 & 1.585 & 2.524\\
3 & 4 & 25 & 2 & 12.5\\
4 & 5 & 1 & 2.322& 0.431\\
5 & 6 & 0 & 2.807 & 0\\
\hline
\end{tabular}
\label{estimated_DCG_table}
\end{center}
\end{table}

Next, a sorted sequence of $m(f_{s}(v), f_{a}(u))$ with length $L = |nbr(v)|$, $u \in nbr(v)$,  is constructed as the estimated ranking, denoted by $B$. Let $B = [1, 2, 4, 5, 6]$. The graded relevance for $B[j]$ depends on the position of $B[j]$ in $A$ and follows the rule: 
\begin{align*}
rel_{B}[j] = \begin{cases} rel_{A}[i], \ \ i\!f \ (\exists i \ : \  A[i] = B[j]) \\
0, \ \ \ \ \ \ \ \ \ otherwise
\end{cases}
\label{}
\end{align*} 
Then $B$'s corresponding $rel_{B}[j]$ and $\frac{rel_{B}[j]}{\log_{2}(j+1)}$ values are shown in Table~\ref{estimated_DCG_table}. Accordingly, $B$'s $DCG_{3} = 16+2.524+12.5 = 31.024$. The ranking similarity between $B$ and $A$ is calculated by the ratio of $B$'s DCG to $A$'s DCG, i.e., $\frac{31.024}{39.595} = 0.784$.

\subsection{Ranking Similarity with Distance-To-Destination
Input Feature}
\label{appendix_similarity_d}

For $\langle f_{s}(v), f_{a}(u) \rangle = \langle d(v, D), d(u, D)\rangle$, we examine if there exists a linear function $m$ that yields high $SIM_{v}(m, Q^{*})$ and $SIM_{G}(m, Q^{*})$ across different network configurations. Let the ranking metric $m$ be $m(f_{s}(v), \\ f_{a}(u)) = -d(u, D)$. (Recall that $Q^{*}(f_{s}(v), f_{a}(u))$ is the cumulative negative length of the shortest path.) 

\begin{figure}[hbt!]
    \centering
    \subfigure[Size 50, Density 3, rnd=19]
    {
        \includegraphics[width=0.22\textwidth]{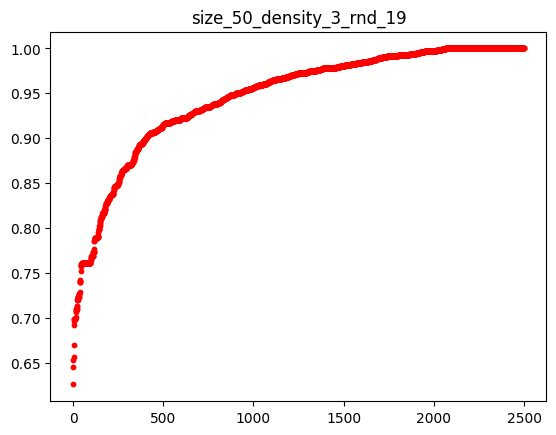}
        \label{d_node_similarity_size50_density3}
    }
    \subfigure[Size 50, Density 5, rnd=19]
    {
        \includegraphics[width=0.22\textwidth]{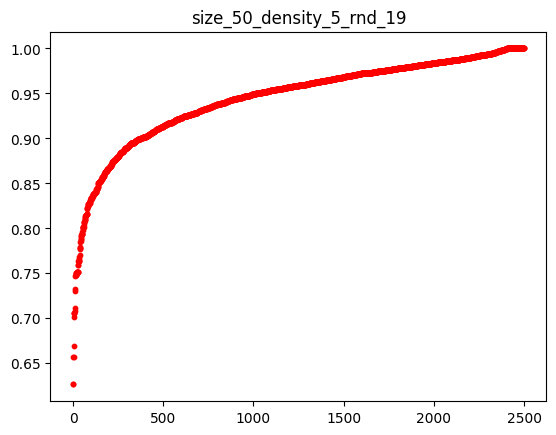}
        \label{d_node_similarity_size50_density5}
    }
    \subfigure[Size 100, Density 3, rnd=48]
    {
        \includegraphics[width=0.22\textwidth]{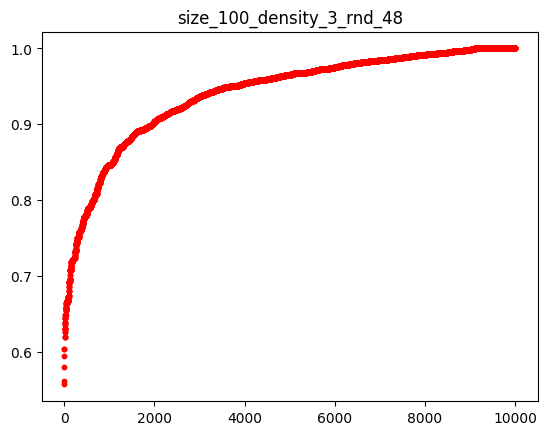}
        \label{d_node_similarity_size100_density3}
    }
    \subfigure[Size 100, Density 5, rnd=48]
    {
        \includegraphics[width=0.22\textwidth]{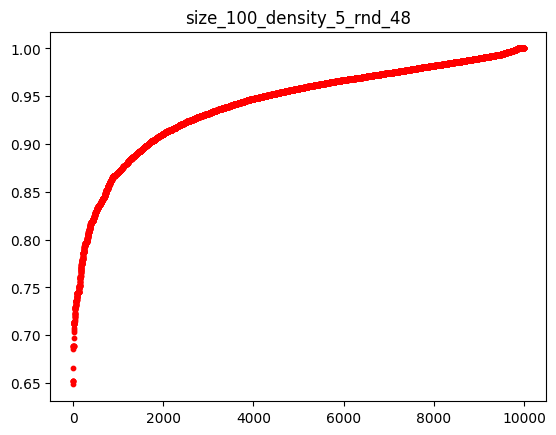}
        \label{d_node_similarity_size100_density5}
    }
    \caption{Distribution of $SIM_{v}(m, Q^{*})$ given a uniform random graph in the Euclidean space with random seed (rnd), where $m$ is the ranking metric for distance-to-destination $d(u ,D)$.}
    \label{d_node_similarity_euclidean}
\end{figure}

In Figure~\ref{d_node_similarity_euclidean}, we plot $SIM_{v}(m, Q^{*})$ for all $v, D \in V$ for a given  random Euclidean graph  with size in \{50, 100\} and density in \{3, 5\}. Each point represents the value of $SIM_{v}(m, Q^{*})$ for a given $v$ and $D$. All $|V|^2$ points are shown in ascending order. The sub-figures in Figure~\ref{d_node_similarity_euclidean} illustrate that at least 75\% of the points have high similarity ($\geq 90\%$) between $m$ and $Q^{*}$. According to Lemma~\ref{cross_node_generalizability}, the distribution of $SIM_{v}(m, Q^{*})$ implies that training samples collected from one (or a few) nodes in this large set can be sufficient to learn a near-optimal routing policy that achieves high accuracy across nodes. On the other hand, using training samples generated from the nodes with relatively low $SIM_{v}(m, Q^{*})$ should be avoided. This observation motivates a knowledge-guide mechanism to carefully choose the data samples used for training.

\begin{figure}[hbt!]
    \centering
    \subfigure[Size 50, Density 3]
    {
        \includegraphics[width=0.22\textwidth]{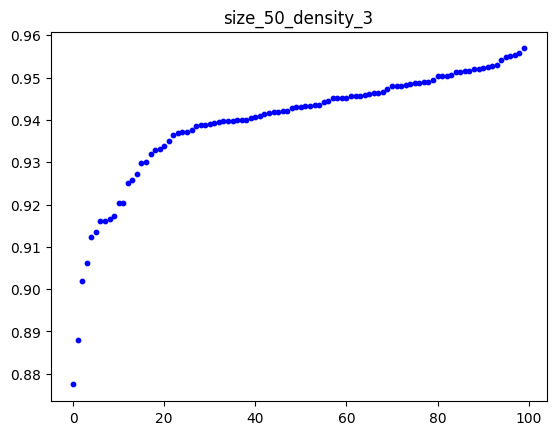}
        \label{d_graph_similarity_size50_density3}
    }
    \subfigure[Size 50, Density 5]
    {
        \includegraphics[width=0.22\textwidth]{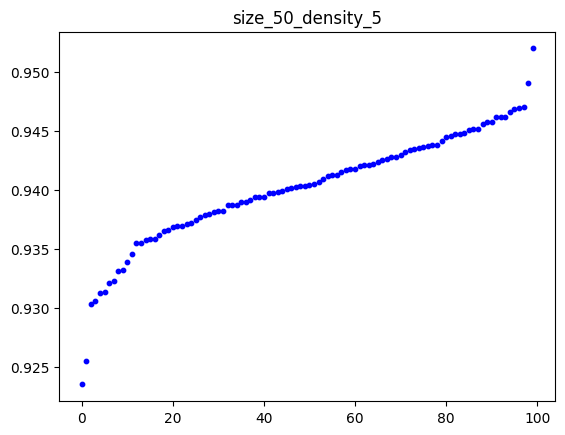}
        \label{d_graph_similarity_size50_density5}
    }
    \subfigure[Size 100, Density 3]
    {
        \includegraphics[width=0.22\textwidth]{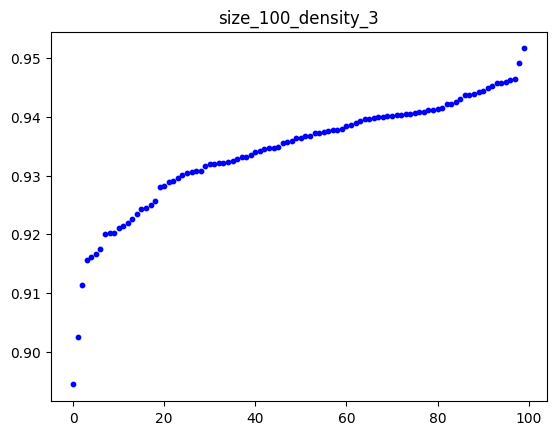}
        \label{d_graph_similarity_size100_density3}
    }
    \subfigure[Size 100, Density 5]
    {
        \includegraphics[width=0.22\textwidth]{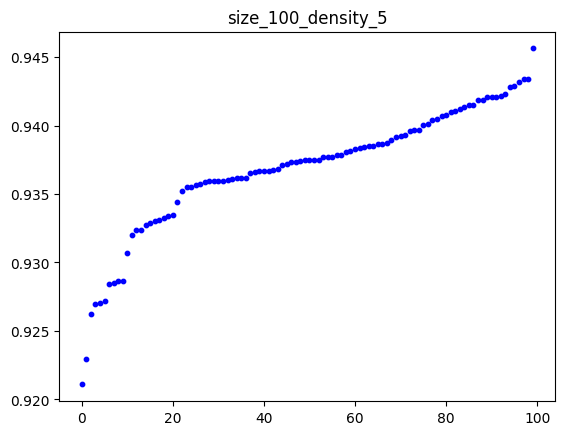}
        \label{d_graph_similarity_size100_density5}
    }
    \caption{Distribution of $SIM_{G}(m, Q^{*})$ across 100 uniform random graphs in the Euclidean space, where $m$ is the ranking metric for distance-to-destination $d(u, D)$.}
    \label{d_graph_similarity_euclidean}
\end{figure}

In Figure~\ref{d_graph_similarity_euclidean}, we show the distribution of $SIM_{G}(m, Q^{*})$ in ascending order for a set of 100 graphs, respectively with size in \{50, 100\} with density in \{3, 5\} in the Euclidean space. Note that in high density Euclidean graphs (say density 5), all 100 graphs have high similarity ($\geq 90\%$) between $m$ and $Q^{*}$, implying that training with samples from almost any high density Euclidean graph  can be sufficient to learn a routing policy with high performance. On the other hand, in low density Euclidean graphs (say density 3), there exist a small set of graphs with slightly low $SIM_{G}(m, Q^{*})$, pointing to the importance of careful selection of seed graph(s) for training. 

In addition, we observe that the upper bound of $SIM_{G}(m, Q^{*})$ decreases as the network size increases. These  results implicitly show that, with respect to the chosen input feature, graphs with smaller size but higher density can be better seeds for learning generalized routing policies.

According to the results in Figure~\ref{d_node_similarity_euclidean} , using distance-to-destination implies the existence of a ranking metric that should with high probability satisfy cross-node generalizability and cross-graph generalizability. The ranking function, $m(f_{s}(v), \\ f_{a}(u)) = -d(u, D)$, or an analogue can be easily learned by a DNN, and then the learned routing policy should achieve high performance across uniform random graphs.

\subsection{Ranking Similarity with Distance-To-Destination and Node Stretch
Input Features}
\label{appendix_similarity_d_sf}

\begin{figure}[hbt!]
    \centering
    \subfigure[Size 50, Density 3, rnd=19]
    {
        \includegraphics[width=0.22\textwidth]{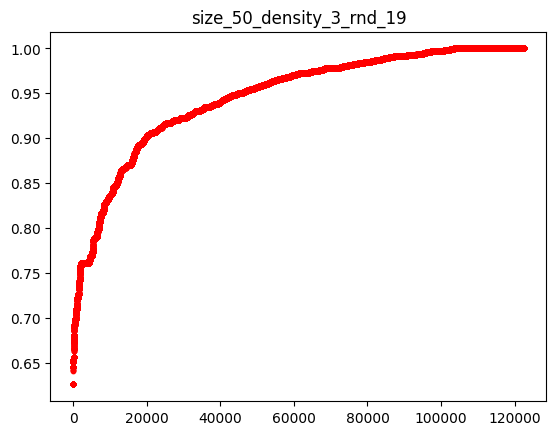}
        \label{d_sf_node_similarity_size50_density3}
    }
    \subfigure[Size 50, Density 5, rnd=19]
    {
        \includegraphics[width=0.22\textwidth]{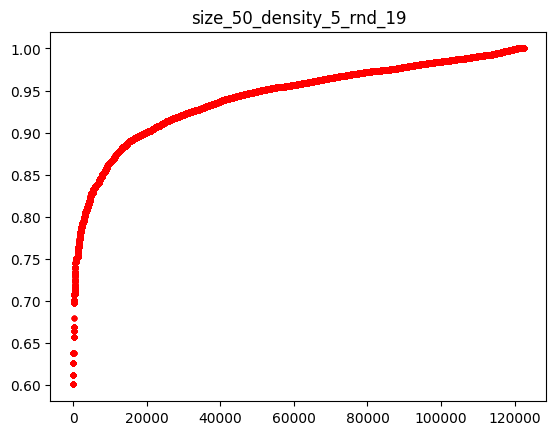}
        \label{d_sf_node_similarity_size50_density5}
    }
    \subfigure[Size 100, Density 3, rnd=48]
    {
        \includegraphics[width=0.22\textwidth]{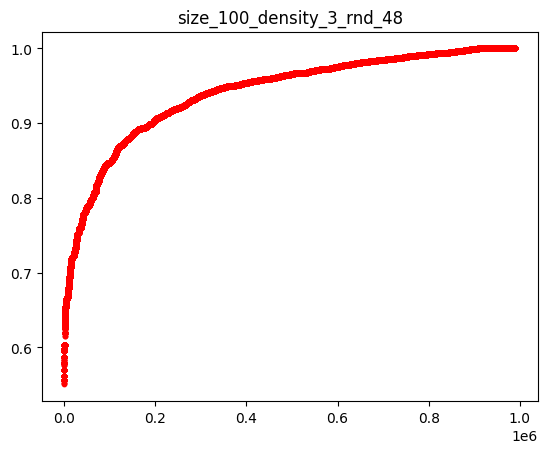}
        \label{d_sf_node_similarity_size100_density3}
    }
    \subfigure[Size 100, Density 5, rnd=48]
    {
        \includegraphics[width=0.22\textwidth]{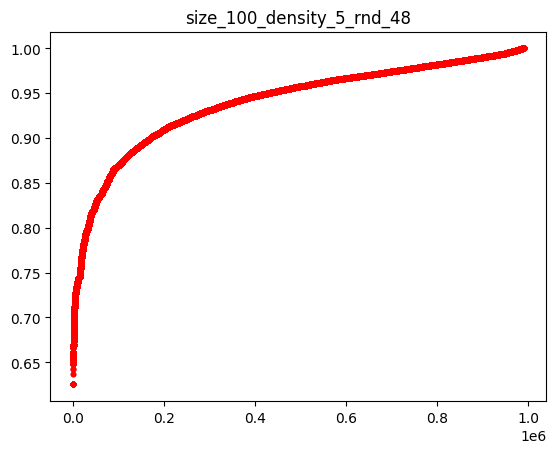}
        \label{d_sf_node_similarity_size100_density5}
    }
    \caption{Distribution of $SIM_{v}(m, Q^{*})$ given a uniform random graph in the Euclidean space with a random seed (rnd), where $m$ is the ranking metric for distance-to-destination $d(u, D)$ and node stretch $ns(O, D, u)$.}
    \label{d_sf_node_similarity_euclidean}
\end{figure}

\begin{figure}[hbt!]
    \centering
    \subfigure[Size 50, Density 3]
    {
        \includegraphics[width=0.22\textwidth]{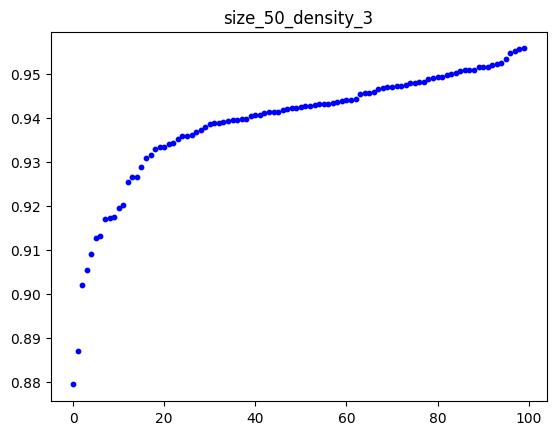}
        \label{d_sf_graph_similarity_size50_density3}
    }
    \subfigure[Size 50, Density 5]
    {
        \includegraphics[width=0.22\textwidth]{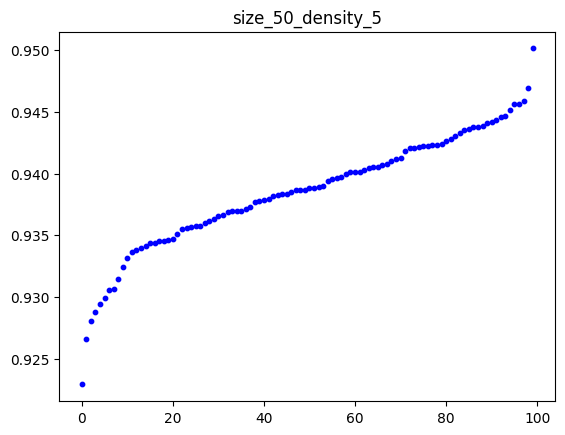}
        \label{d_sf_graph_similarity_size50_density5}
    }
    \subfigure[Size 100, Density 3]
    {
        \includegraphics[width=0.22\textwidth]{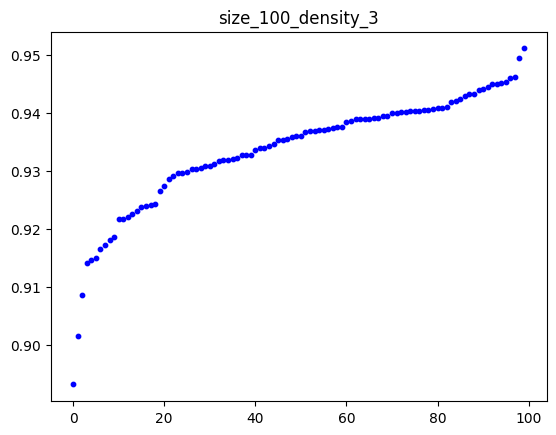}
        \label{d_sf_graph_similarity_size100_density3}
    }
    \subfigure[Size 100, Density 5]
    {
        \includegraphics[width=0.22\textwidth]{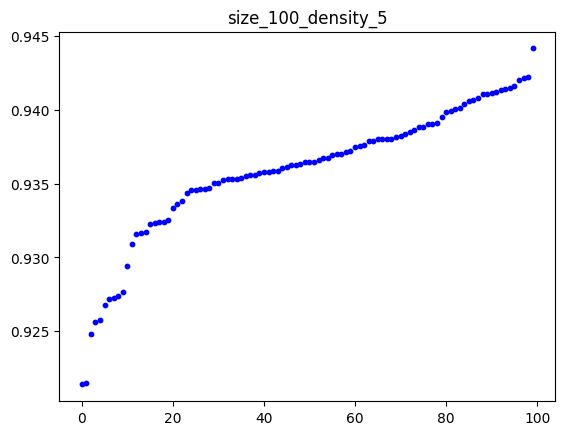}
        \label{d_sf_graph_similarity_size100_density5}
    }
    \caption{Distribution of $SIM_{G}(m, Q^{*})$ across 100 uniform random graphs in the Euclidean space, where $m$ is the ranking metric for distance-to-destination $d(u, D)$ and node stretch $ns(O, D, u)$.}
    \label{d_sf_graph_similarity_euclidean}
\end{figure}

We examine if there exists a linear function $m$ that yields high $SIM_{v}(m, Q^{*})$ and $SIM_{G}(m, Q^{*})$ across different network configurations, for $\langle f_{s}(v), f_{a}(u) \rangle = \langle d(v, D), ns(O, D, v), d(u, D), ns(O, D, u)\rangle$.  Let the ranking metric $m$ as $m(f_{s}(v), \\ f_{a}(u)) = -0.875 d(u, D) - 0.277 ns(O, D, u)$ in the Euclidean graphs. Note that $f_{s}(v)$ does not affect the sorting of $m(f_{s}(v), f_{a}(u))$. For convenience, we assign zero weight for $d(v, D)$ and $ns(O, D, v)$ in $m$.

In Figures~\ref{d_sf_node_similarity_euclidean} and~\ref{d_sf_graph_similarity_euclidean}, we apply the same network configurations in the Euclidean graphs as in Figures~\ref{d_node_similarity_euclidean} and~\ref{d_graph_similarity_euclidean} to plot the distribution of $SIM_{v}(m, Q^{*})$ and $SIM_{G}(m, Q^{*})$ for the proposed choice of $m$. Note that because the node stretch relies on $v, O, D \in V$, there are $|V|^3$ points plotted in Figure~\ref{d_sf_node_similarity_euclidean} with ascending order. The sub-figures in Figure~\ref{d_sf_node_similarity_euclidean} demonstrate that at least 80\% of the points have similarity above 90\%, which is higher than the corresponding percentage of points in Figure~\ref{d_node_similarity_euclidean}. These observations indicate that using both distance-to-destination and node stretch assures the existence of ranking metric, and in turn implies learnability of a DNN that with high probability achieves even better cross-node generalizability, compared to the one using only Euclidean distance in the input features.

Figure~\ref{d_sf_graph_similarity_euclidean} shows a similar distribution of $SIM_{G}(m, Q^{*})$ to Figure~\ref{d_graph_similarity_euclidean}, respectively. In order to achieve cross-graph generalizability, we need to carefully select a seed graph for training, especially from graphs with moderate size and high density. Also, instead of using all the nodes to generate data samples, using a subset of nodes that avoid relatively low $SIM_{v}(m, Q^{*})$ further improves the cross-node generalizability.

\section{Reinforcement Learning Algorithm for APNSP}
\label{appendix_RL_algo}
More specifically, in Algorithm~\ref{RL_ALGO}, Lines 2 to 20 outline the sample selection and training procedure for each episode. The for-loop from Lines 4 to 7 determines the set of chosen shortest paths and the associated nodes for subsampling, where the shortest paths are predicted based on the current DNN estimation of $Q(v, u)$ for a routing node $v$ and its neighbors $u \in nbr(v)$. The neighbor $u$ is chosen to be the next routing node until the destination is reached or none of the neighbors remain unvisited. Note that each node can be only visited once in a round of path exploration. It is possible to output a path $p$ without the destination $D$, which  can still be used for subsampling in this episode. The for-loop in Lines 9 to 16 generates the training data samples from nodes shown in the chosen paths, where the data labels $Y$, i.e., target $Q$-values that we train the DNN to fit for improving the estimation accuracy of optimal $Q$-values, are given by: 
\begin{align}
    Q^{target}(s, a) = r(s, a) \! + \! \gamma \max_{a'}{Q(s', a')}.
\label{Q_estimate}
\end{align}
Next, based on the collected dataset, in Lines 17 to 19, the DNN is trained for a fixed number of iterations to minimize the following loss function:
\begin{align}\label{fit}
    \min_{\theta}~~\sum_{\langle X, Y\rangle} \|H_{\theta}(f_{s}(v),f_{a}(u)) - Q^{target}(v,u)\|^2.
\end{align}
Since the target $Q$-values approach the optimal $Q$-values as the number of training episodes increases, minimizing Equation~\ref{fit} will eventually lead to a learned model that nearly matches the supervised learning in Equation~\ref{supervised}.

\begin{algorithm}
\SetAlgoLined
Input: $nn$: randomly initialized DNN; $G^{*}$: seed graph; $\Phi$: set of chosen sources; $D$: chosen destination

\For{$episode = 1 ... EpiNum$}{
$V_{T}$ := \{\}\;
\For{$O \in \Phi$}{
  Use $nn$ to predict a shortest path $p$ for $(O, D)$ in $G^{*}$\;
  $V_{T}$ := $V_{T} \cup \{v| v \in p\}$\;
}
$X$:=[], $Y$:=[], 
$i := 0$\;
\For{$v \in V_{T}$}{
  \For{$u \in nbr(v)$}{
    $X[i] := \langle f_{s}(v),f_{a}(u)\rangle$\;
    Estimate $Q(v, u)$ using Equation~\ref{Q_estimate}\;
    $Y[i] := Q(v, u)$\;
    $i := i+1$\;
    }
}
\For{$iter = 1 ... IterNum$}{
  Train $nn$ with $\langle X, Y \rangle$ based on Equation~\ref{fit}\;
}
}

\Return $nn$\;
\caption{RL-APNSP-ALGO}
\label{RL_ALGO}
\end{algorithm}

\section{Extended Experimental Results}
\label{appendix_results}
\begin{table}[htbp]
\caption{Simulation Parameters}
\begin{center}
\begin{tabular}{c c c}
\hline
\textbf{Symbol}& \textbf{Meaning} & \textbf{Value} \\
\hline

$N_{train}$ & size of seed graph & 50 \\
$\rho_{train}$ & density of seed graph & 5 \\ 
$N_{test}$ & sizes of tested graphs & 27, 64, 125, 216 \\
$\rho_{test}$ & densities of tested graphs & 2, 3, 4, 5 \\
$R$ & communication radius & 1000 \\
$\Omega=I+J$ & \# of input features & 2,4 \\
$K$ & \# of hidden layers & 2 \\
$N_{e}[]$ & \# of neurons in each hidden layer  & $[50\Omega, \Omega]$ \\
$\epsilon$ & margin for shortest paths prediction & 0.05 \\
$\phi$ & \# of nodes for subsampling & 3 \\
$\gamma$ & discount factor & 1 \\
$IterNum_{S}$ & \# of iterations in supervised learning & 5000 \\ 
$IterNum_{RL}$ & \# of iterations in RL & 1000 \\ 
$EpiNum$ & \# of episodes in RL & 20 \\ 
\hline
\end{tabular}
\label{SimulationParams}
\end{center}
\end{table}
\subsection{Zero-shot Generalization over Hyperbolic Graphs}

{\bf Uniform Random Graphs in Hyperbolic Plane.}
For a graph $G$ in a 2-dimensional hyperbolic plane, all nodes in $V$ are distributed in a disk of radius $R$. Each node $v$ thus has hyperbolic polar coordinates $(r_{v}, \theta_{v})$ with $r_{v} \in [0, R]$ and $\theta_{v} \in [0, 2\pi)$. Let $\delta$ be the average node degree. Let $n$ and $-\alpha$ denote the number of nodes in $V$ and the negative curvature, respectively. All nodes have their radial coordinates sampled according to the probability distribution $p(r) = \alpha \frac{\sinh{(\alpha r)}}{\cosh{(\alpha R)} -1}$ and their angular coordinates sampled uniformly at random from the interval $[0, 2\pi)$. 
It is well known that such uniform random graphs in the hyperbolic plane yield a power-law distribution for the node degrees \cite{aldecoa2015hyperbolic}.

\begin{table}[htbp]
\caption{Simulation Parameters for Hyperbolic Graphs}
\begin{center}
\begin{tabular}{c c c}
\hline
\textbf{Symbol}& \textbf{Meaning} & \textbf{Value} \\
\hline
$\delta_{test}$ & avg. node degree of tested graphs & 1, 2, 3, 4 \\
 & in hyperbolic space & \\
$-\alpha$ & negative curvature in hyperbolic space & -0.6 \\
\hline
\end{tabular}
\label{SimulationParams_hyperbolic}
\end{center}
\end{table}

\begin{figure*}[hbt!]
    \centering
    \subfigure
    {
    \includegraphics[width=0.6\textwidth]{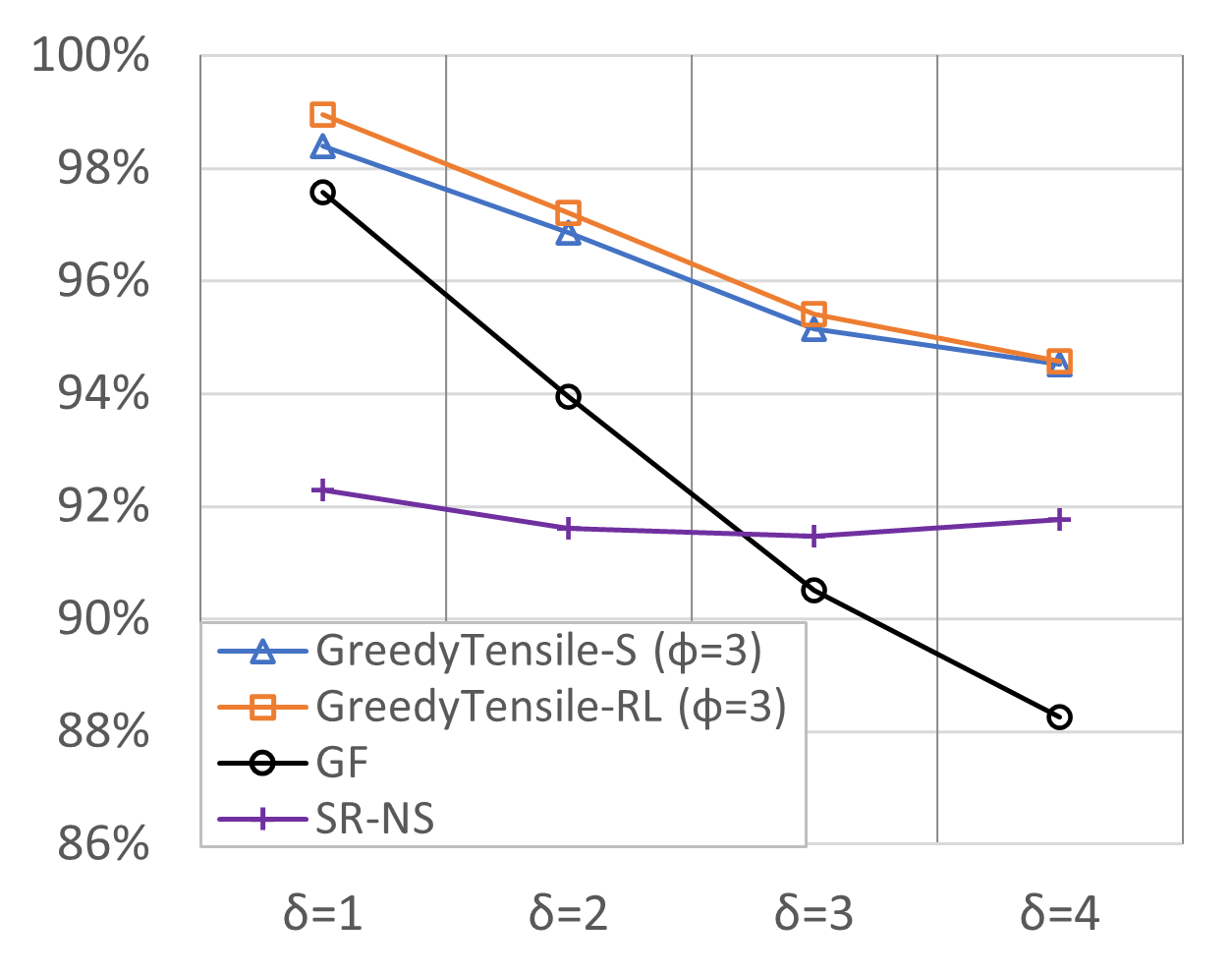}
        \label{testing_hyperbolic}
    }
    \caption{Average APNSP prediction accuracy across graph sizes with various $\delta$ in hyperbolic space for \GreedyTensile policies.}
    \label{Testing_Performance_ed_sf_hyperbolic}
\end{figure*}

\noindent
{\bf Evaluation Results.} In hyperbolic space, we apply the same network configuration as in Table~\ref{SimulationParams}. The parameters specific to the hyperbolic graphs are shown in Table~\ref{SimulationParams_hyperbolic}. 

For the \GreedyTensile DNNs, we plot in Figure~\ref{Testing_Performance_ed_sf_hyperbolic} the respective average prediction accuracies across graph sizes in \{27, 64, 125, 216\} with average node degree in \{1, 2, 3, 4\} in hyperbolic space. 
Both the GreedyTensile-S and GreedyTensile-RL policy shows comparable performance across graphs with different $\delta$.
Both are more accurate than GF (by up to 6.27\%) and SR-NS (by up to 6.11\%). (Whereas SR-NS outperforms GF in high-degree graphs ($\delta \geq 3$) but incurs a significant performance gap to other policies in low-degree graphs ($\delta \leq 2$)).

\subsection{Ablation Study}
\label{ablation_study}

\begin{figure}[hbt!]
    \centering
    \subfigure[GreedyTensile-S]
    {
        \includegraphics[width=0.35\textwidth]{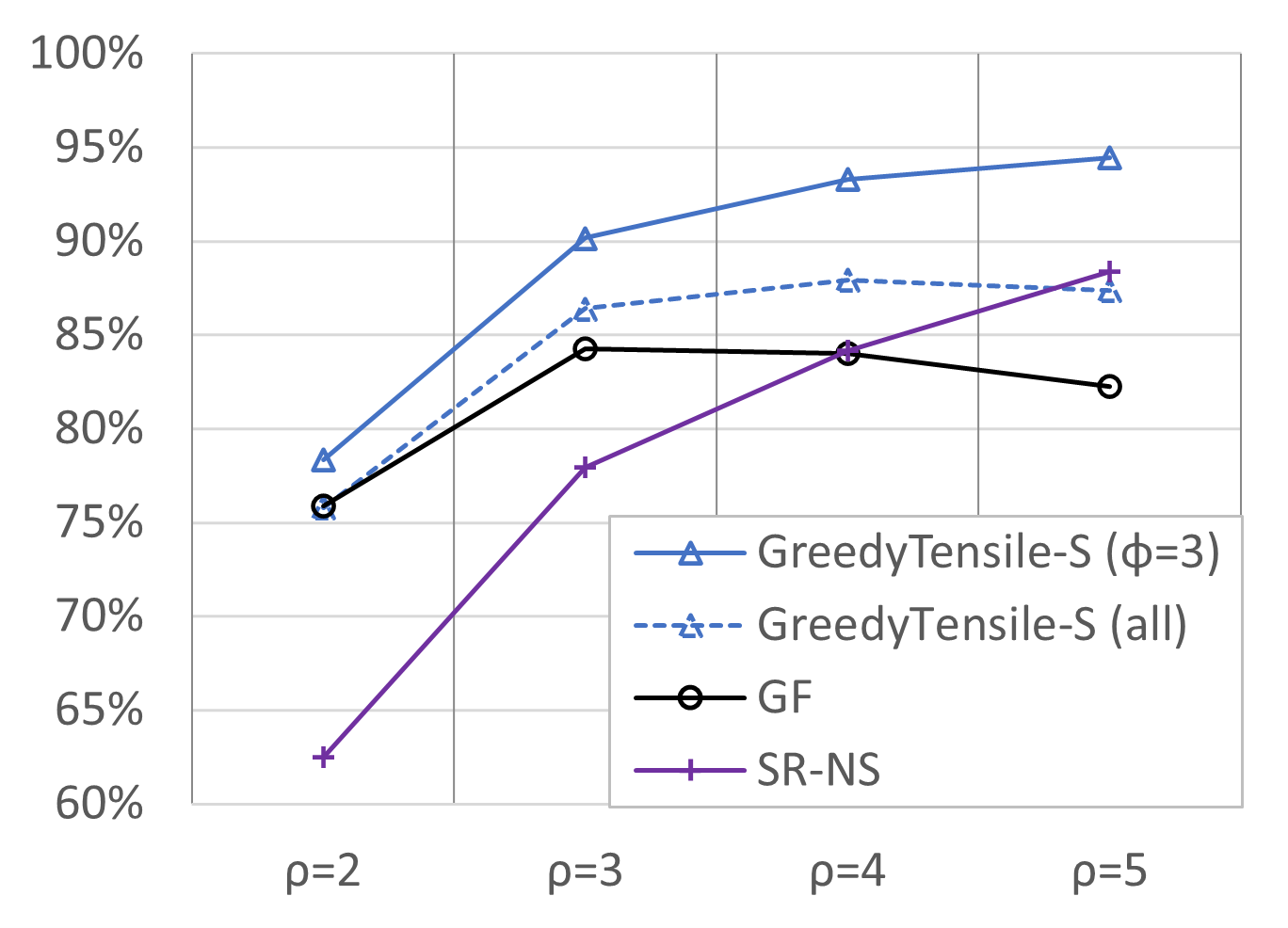}
        \label{ablation_euclidean_GT_S}
    }
    \subfigure[GreedyTensile-RL]
    {
        \includegraphics[width=0.35\textwidth]{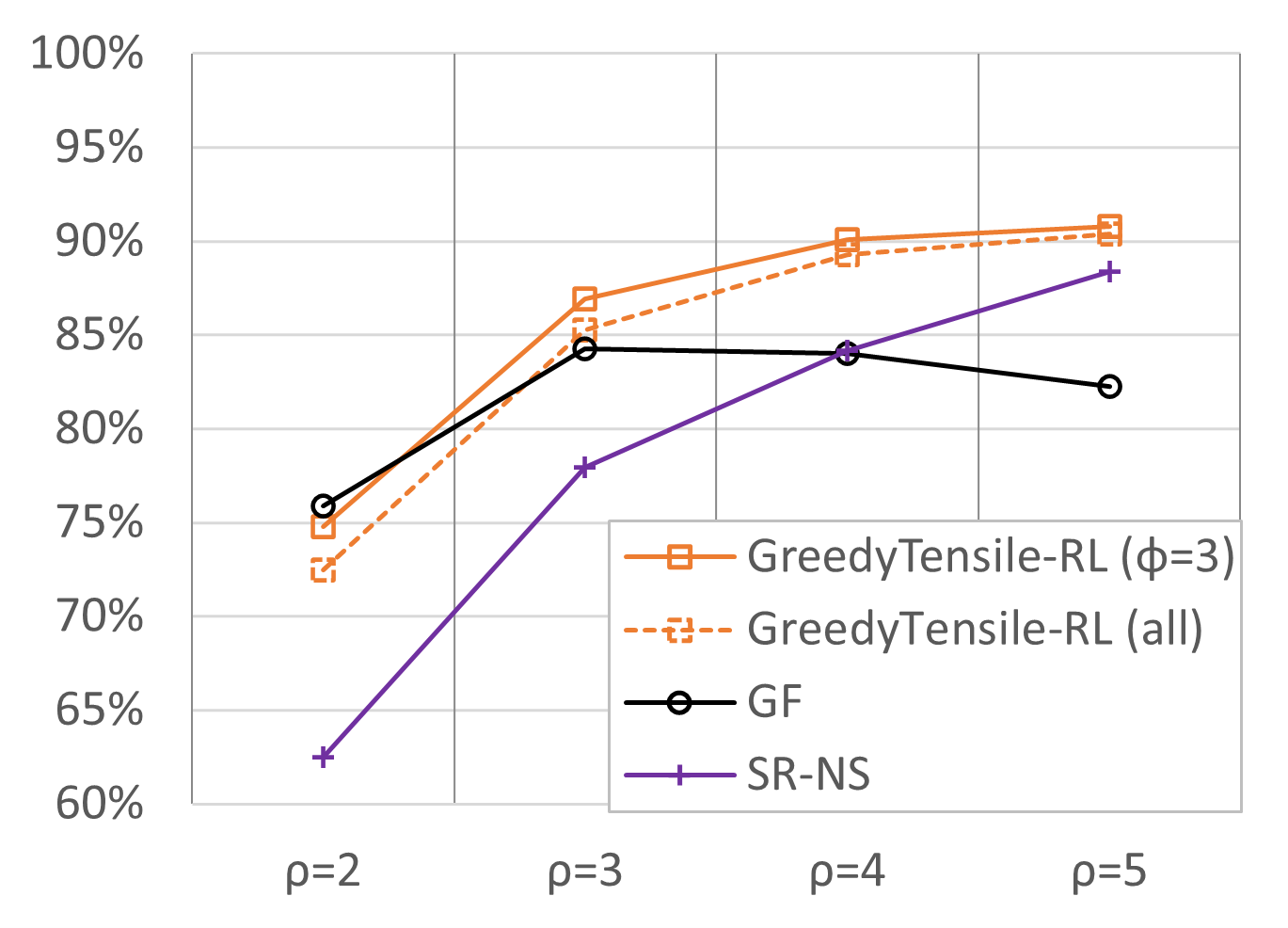}
        \label{ablation_euclidean_GT_RL}
    }
    \caption{Average APNSP prediction accuracy across graph sizes with various $\rho$ in Eucldiean space for \GreedyTensile policies with and without graph subsampling.}
\label{ablation_euclidean}
\end{figure}

To evaluate the effect of graph subsampling in the performance of DNN models, we compare \GreedyTensile policies learned from training samples derived from
$\phi=3$ nodes and all nodes in terms of APNSP prediction accuracy. Figure~\ref{ablation_euclidean} shows the average prediction accuracies across graph sizes in \{27, 64, 125, 216\} with density in \{2, 3, 4, 5\} for GreedyTensile-S and GreedyTensile-RL DNNs. The gap between DNNs with graph subsampling ($\phi=3$) and without graph subsampling ($\phi=n$, all) implies that the selection of data samples not only reduces the sampling and training complexity, but also improves the performance of learned models.

\section{Symbolic Interpretability of Learned Model}
\label{interpretability}

\begin{figure}[hbt!]
    \centering
    \subfigure[\em \GreedyTensile DNN]
    {
        \includegraphics[width=0.35\textwidth]{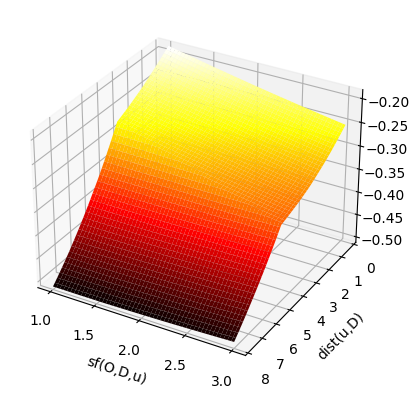}
        \label{model_euclidean_v2d_4}
    }
    \subfigure[{\em Symbolic Approximation}]
    {
        \includegraphics[width=0.35\textwidth]{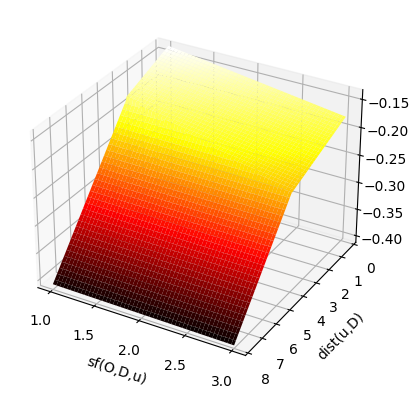}
        \label{model_two_linear_actions}
    }
    \caption{The shape of ranking metrics of the \GreedyTensile DNN and its two-linear action Symbolic Approximation policy, given $s(O, D, v) = 1.2$ and $d(v, D)=4$. The x and y axes represent $s(O, D, u)$ and $d(u, D)$, and the z axis is the ranking metric for routing.}
    \label{model_euclidean}
\end{figure}

In this section, we symbolically interpret the learned model for \GreedyTensile routing, achieving a two orders of magnitude reduction in its operational complexity. Figure~\ref{model_euclidean} plots the output of its DNN, towards explaining the learned policy\footnote{Since \GreedyTensile models in Euclidean and hyperbolic spaces have a similar shape, we only visualize the learned model in the Euclidean space here.}. 
Since $d(v, D)$ and $s(O, D, v)$ stay unchanged for a fixed routing node $v$ at a given time, we plot the shape of the ranking metric (z-axis) of the learned DNN according to varying $s(O, D, v)$ (x-axis) and $d(u, D)$ (y-axis) in the figure. 

Figure~\ref{model_euclidean_v2d_4} shows that the \GreedyTensile DNN has two planes separated by a transition boundary.
The two planes can be respectively approximated by two different linear functions. The first function (for the upper plane) prefers both smaller $s(O, D, u)$ and $d(u, D)$. The second (for the lower plane) significantly prioritizes smaller $d(u, D)$. We find that the two functions that approximate the \GreedyTensile DNN can be symbolically represented by a guarded command:

\vspace{-2mm}
\begin{align}
z = \begin{cases} -0.01d(v, D) - 0.02ns(O, D, u) - 0.01d(u, D) - 0.06, \\ \ \ \ \ i\!f \ d(u, D) <  1.02d(v, D) + 0.57ns(O, D, u) - 0.69 
\\
0.03d(v, D) - 0.04d(u, D) - 0.15, \ \ otherwise.
\end{cases}
\label{guarded_command}
\end{align}
\vspace{-2mm}

\begin{figure}[thb!]
    \vspace{-2mm}
    \centering
     \subfigure
     {
        \includegraphics[width=0.5\textwidth]{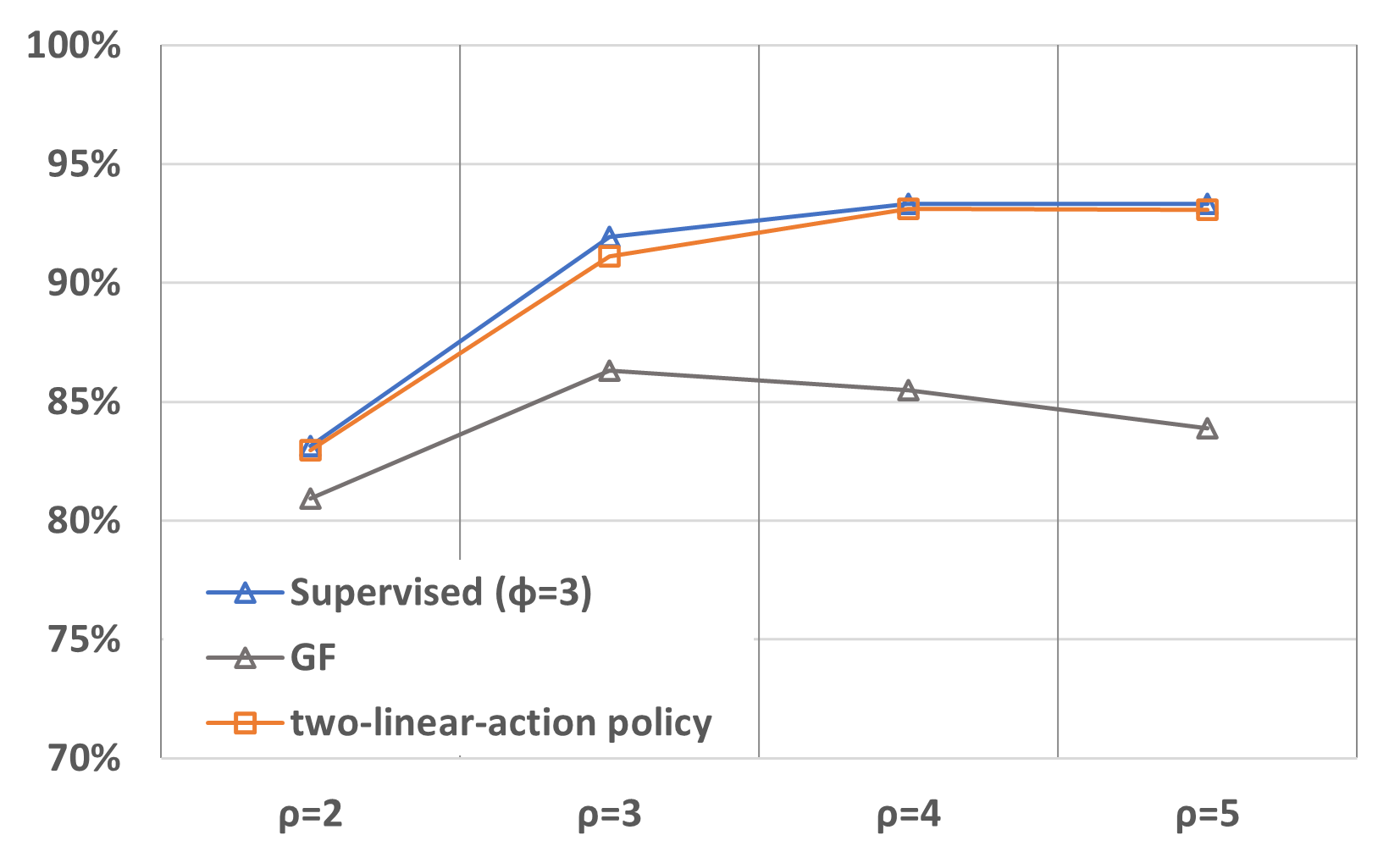}
    }
    \vspace{-2mm}
    \caption{APNSP prediction accuracy for the two-linear-action policy in Equation~\ref{guarded_command} versus the policies of \GreedyTensile (Supervised ($\phi=3$)) and greedy forwarding (GF) over graphs in the Euclidean space with size 50 and density in $\{2, 3, 4, 5\}$.}
    \label{two_linear_action_evaluation}
\end{figure}

The weights of the guarded command are calculated using linear regression. Its first action assigns dominant weights to both $s(O, D, u)$ and $d(u, D)$, while its second action gives zero weight to $ns(O, D, u)$.
The shape of ranking metrics of the guarded command given $d(v, D) = 4$ is shown in Figure~\ref{model_two_linear_actions}, which has a two-plane surface similar to Figure~\ref{model_euclidean_v2d_4}. 
Figure~\ref{two_linear_action_evaluation} shows that the accuracy of the two-linear-action policy using Equation~\ref{guarded_command} is close to that of \GreedyTensile DNN.

The simplified two-linear-action policy also has substantially reduced operation complexity. Whereas the \GreedyTensile DNN requires at least $\Omega \times N_{e}[1] \times N_{e}[2] (= 4*200*4)$ multiplications to output the $Q$-value for a given (state, action) pair, where $\Omega$ represents the number of input features of the DNN and $N_{e}[i]$ denotes the number of neurons in the $i$-th hidden layer, the two-linear-action policy needs less than ten multiplications.

\end{document}